\documentclass[10pt,twocolumn,letterpaper]{article}

\usepackage[pagenumbers]{cvpr}   %
\usepackage[T1]{fontenc}

\usepackage{cuted}
\usepackage{caption}
\usepackage{graphicx}
\usepackage{amsmath}
\usepackage{amssymb}
\usepackage{booktabs}
\usepackage{multirow}
\usepackage{xcolor}
\usepackage{listings}
\usepackage{siunitx}

\newcommand{\ours}{Scribe3D}

\newcommand{\detourref}{\cref{sec:detour}}
\newcommand{\promptsref}{\cref{app:prompts}}
\newcommand{\fidelityref}{\cref{sec:fidelity}}   %
\newcommand{\editingref}{\cref{sec:editing}}
\newcommand{\dropoutref}{\cref{sec:ablation}}
\newif\ifsupp\suppfalse   %

\definecolor{cvprblue}{rgb}{0.21,0.49,0.74}
\usepackage[pagebackref,breaklinks,colorlinks,allcolors=cvprblue]{hyperref}

\def\confName{arXiv}
\def\confYear{2026}

\title{Scribe3D: Turning VR Sketch Coordinates into Registered and Editable\\3D Part Programs without Training or 2D Detours}

\author{Yichen Yu \qquad Qiao Jin\\
North Carolina State University\\
{\tt\small \{yyu55, qjin4\}@ncsu.edu}
}

\begin{document}
\twocolumn[{%
\renewcommand\twocolumn[1][]{#1}%
\maketitle
\begin{center}
  \centering
  \includegraphics[width=\linewidth]{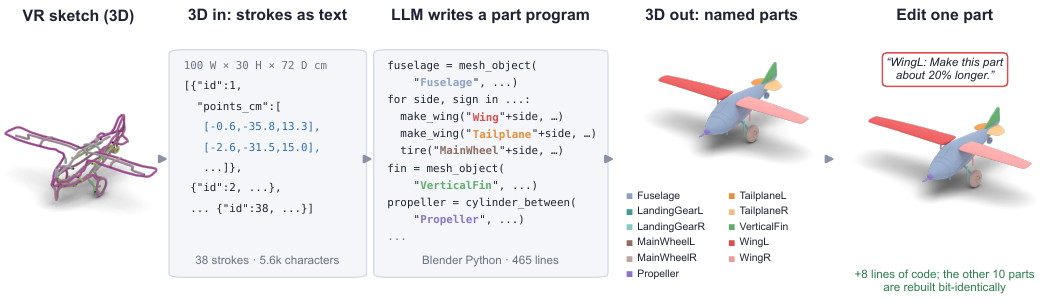}
  \captionof{figure}{\textbf{\ours{}.} A hand-drawn VR sketch (left) is written down as text: its 3D stroke coordinates in a canonical frame. A general-purpose LLM writes a Blender program that builds the object from named parts. A text edit addressed to one part changes only that part's code: here the left wing becomes 20\% longer while the other ten parts are rebuilt bit-identically (right; same scale and part colors as before the edit).}
  \label{fig:teaser}
\end{center}
\vspace{1em}
}]
\begin{abstract}
VR sketches are three-dimensional, yet training-free routes from a sketch to a 3D model typically pass through 2D: the sketch is rendered to images and lifted back by an image or image-to-3D model. We show that a minimal representation lets a general-purpose model stay in 3D instead: the 3D strokes enter a large language model (LLM) as coordinates in a normalized metric frame (line drawings of the strokes are attached by default but can be dropped without loss), and the LLM returns a Blender program that builds the object as separately named 3D parts in that same frame, so the result appears where, and at the size, the user drew it, can be checked against the strokes without ground truth, and can be edited one part at a time; these properties follow from the shared frame and the program's named parts rather than from a registration or segmentation step. Our pipeline built on this representation, \ours{}, trains nothing. On all 400 hand-drawn sketches of the VRSketch2Shape benchmark, \ours{} obtains a higher F-score than the trained state of the art (71.3 vs.\ 69.8, $p=0.005$), does not differ significantly in the official Chamfer distance or in F-score after ICP alignment, and is worse in Chamfer distance after ICP alignment and in normal consistency. Switching the prompt's shape-fidelity instruction from balanced to strict, a variant we tested after seeing the balanced results, raises the F-score to 74.1 vs.\ 69.5 for the trained model on the 320 sketches not used to develop it ($p<10^{-8}$). The approach is not tied to one LLM: Claude Fable 5.2 and Gemini 4 Argon also reach the trained model's level on 60 sketches, whereas the strongest open-weight model we tried does not. Nor is it tied to our pipeline: a re-implementation of LL3M's multi-agent loop, given the same LLM and coordinates, reaches a level comparable to ours (72.2 vs.\ 70.6 on 40 sketches), so the representation, more than our prompt, carries the result. On synthetic sketches of four categories the trained model has not seen, \ours{} scores higher than it on 20 of 24; on people and plants, far from the benchmark's rigid objects, it builds figures in the sketched pose and plants leaf by leaf, and scores higher on 19 of 20 against the original meshes. The 3D channel carries the signal: giving the same LLM rendered views of the strokes instead of their coordinates costs 7.8 F-score points and leaves \ours{} significantly below the trained model, while coordinates without views match coordinates plus views; a complete 2D detour built from public models, which renders the sketch, generates an image and lifts it to 3D with Hunyuan3D-2, reaches an F-score of only 43.3 on 100 sketches, against 71.3 for \ours{} with coordinates but no views. Because the program is written in the sketch's frame, the result needs no registration: the drawn strokes lie a median 0.54\% of the sketch size from our surface, closer than the trained model's output even after fitting it to the sketch's bounding box (370 of 400 sketches). Because the output is a program of named parts, a text edit addressed to one part changes only that part: across 59 edits, all 499 untouched parts were rebuilt bit-identically, whereas re-sketching with the trained model moves on average 23\% (same noise seed; 12\% after alignment) of the rest of the surface, ignored edits included. Two VLMs judge 44 and 45 of the 59 edits as done as asked.
\end{abstract}

\section{Introduction}
\label{sec:intro}

Sketching in virtual reality lets a designer draw directly in 3D: strokes have depth, the drawing can be inspected from any side, and the result is a set of 3D polylines rather than a projection. Turning such a sketch into a usable 3D shape is the goal of a line of work on VR sketch-based retrieval and modeling~\cite{luo2021fine,luo2022structure,luo2023vrsketch,chen2026vrsketch2shape}. The current state of the art, VRSketch2Shape~\cite{chen2026vrsketch2shape}, trains a sketch-conditioned latent diffusion model and reaches an F-score of 69.8 on a benchmark of 400 hand-drawn VR sketches.

Existing routes from a 3D sketch to a 3D shape fall into two groups. \emph{Trained 3D-to-3D models}~\cite{luo2023vrsketch,chen2024deep3dvrsketch,chen2026vrsketch2shape} need paired sketch--shape data, which is scarce for VR sketches and limited to a few categories; they output a single surface without parts, generated in a canonical frame rather than where the user drew, or, for CAD parts drawn with construction lines, a trained model recovers a CAD program~\cite{li2026cadrawer}. \emph{Training-free routes} instead take a detour through 2D: the sketch is rendered to images, an image model interprets them, and an image-to-3D generator lifts the result back to 3D~\cite{huang2026handmade}; LLM-based modeling systems likewise take 2D sketches or images as visual input~\cite{noor2026nova3d,lu2025ll3m}. The detour discards exactly what makes a VR sketch valuable, its depth and its placement in space, and then has to recover it.

We ask whether a VR sketch can reach a general-purpose model as 3D geometry, rather than as images, and come back as a 3D model in the same place. Large language models have become capable 3D programmers that write 3D modeling scripts~\cite{sun20233dgpt,hu2024scenecraft}, including ones that build objects from named parts~\cite{lu2025ll3m,noor2026nova3d}, and they can read geometry serialized as text~\cite{wang2024llamamesh,qiu2025sgpbench}. We study the minimal representation that connects the two (\cref{fig:teaser}): the sketch's strokes written down as coordinates in a normalized metric frame, the same frame in which the LLM writes a Blender program that builds the object as separately named 3D parts. Nothing is trained, and no image is generated. Three properties then follow from the shared frame and the part program rather than from a registration or segmentation step: the generated object is registered to the sketch; it can be compared with the strokes without ground truth; and, being a program of named parts, it can be repaired or edited one part at a time. \ours{} is our pipeline built on this representation; a second pipeline of different design, given the same coordinates, reaches a comparable level.

Our central finding is that this 3D input representation, more than a particular prompt or pipeline, lets general-purpose LLMs produce, without task-specific training, 3D models that compete with a trained model and, unlike its output, are registered to the sketch, verifiable against it and editable part by part. On the VRSketch2Shape benchmark:
\begin{itemize}
  \item \textbf{The representation carries the result.} Given the same LLM and prompt, rendered views of the strokes instead of their coordinates lose 7.8 F-score points and fall significantly below the trained model, while coordinates without views match coordinates plus views. A complete 2D detour, rendering the sketch, generating an image from it and lifting the image with an image-to-3D model, reaches 43.3 against 71.3 for coordinates without views on 100 sketches. Our prompt adds to the coordinates rather than replacing them: told only to write a Blender script, the same LLM reaches 65.9 against 70.6 on the same 40 sketches, once the ground plane it adds is removed. A re-implementation of LL3M's multi-agent loop with the same LLM and coordinates reaches a level comparable to ours (72.2), at 2.9$\times$ the cost, and Claude Fable 5.2 and Gemini 4 Argon in our pipeline also reach the trained model's level: the result rests on the coordinates in the program's frame, not on our prompt, pipeline or LLM. Removing the category name does not lower the score (\cref{sec:ablation,sec:openllm}).
  \item \textbf{Training-free, on par with the trained state of the art.} On all 400 hand-drawn test sketches, \ours{} obtains 71.3 vs.\ 69.8 ($p=0.005$) with no training and no benchmark-specific prompt tuning. It is significantly better on airplanes, chairs and tables, not significantly different in the official Chamfer distance and in F-score after ICP alignment, and worse in Chamfer distance after ICP alignment and in normal consistency (\cref{sec:main,sec:more_metrics}). Changing one sentence of the prompt to ask for strict fidelity to the strokes raises the F-score further, to 74.1 vs.\ 69.5 on the 320 sketches not used to develop it ($p<10^{-8}$; \cref{sec:fidelity}). On synthetic sketches of four categories the trained model has not seen, \ours{} scores higher on 20 of 24 (\cref{sec:ood}).
  \item \textbf{The result stays where the user drew.} Without any registration, the drawn strokes lie a median 0.54\% of the sketch size from our surfaces (96.8\% of stroke points within 2\%). The trained model generates in a canonical frame and, even after fitting it to the sketch's bounding box, is farther from the strokes on 370 of 400 sketches. Rotated sketches yield rotated outputs (\ang{45}: 64.2 vs.\ 43.7 for the trained model, which generates in its canonical orientation). Comparing output and strokes, without ground truth, points to the parts to repair; repair raises the F-score on held-out sketches with our acceptance score and with two simpler ones (\cref{sec:registration,sec:serialization,sec:verify_results}).
  \item \textbf{Edits stay local, and we measure it.} Across 59 part-level edits, all 499 parts that an edit did not target were rebuilt bit-identically, and so were all untouched parts over chains of three successive edits. Achieving 30 of the 32 rescaling edits with the trained model by re-sketching the part moves on average 23\% (same noise seed; 12\% after alignment) of the surface outside the target, unchanged outputs included. Scaling a part of a perfectly segmented mesh matches this locality, but such a transform cannot restyle a part or add details (\cref{sec:editing}).
\end{itemize}
We also report what does not work. Cabinets lose largely because the programs build interiors (drawer boxes, shelves) that a solid ground truth does not have; the strict instruction removes some of them, and the gap is then no longer significant (\cref{sec:cabinets,sec:fidelity}); individual results vary between runs; and each shape costs about \$0.31 in API fees and one minute of time. The prompts are given verbatim in \cref{app:prompts}.

\section{Related Work}
\label{sec:related}

\paragraph{3D sketches to 3D shapes.}
Early work on VR sketches retrieves existing shapes~\cite{luo2020towards,luo2021fine,luo2022structure,halder2025fav3r}, also in combination with speech~\cite{giunchi2021mixing}. Generative approaches train conditional models on paired sketches and shapes: Luo \etal~\cite{luo2023vrsketch} generate multiple shapes that respect the sketch structure, Deep3DVRSketch~\cite{chen2024deep3dvrsketch} fine-tunes a sketch encoder for a pre-trained diffusion model, VRSketch2Gaussian~\cite{gu2025vrsketch2gaussian} produces 3D Gaussians from a sketch and text, and VRSketch2Shape~\cite{chen2026vrsketch2shape} encodes the stroke sequence with an order-aware transformer and conditions SDFusion~\cite{cheng2023sdfusion} on it, trained on 20k synthetic and 900 hand-drawn pairs. All of these are trained per dataset and output a single implicit surface, mesh or Gaussian set without named parts. CADrawer~\cite{li2026cadrawer} is closest to us in producing a program from 3D strokes: it reconstructs a CAD program from 3D sketches with dense construction lines by predicting modeling operations autoregressively, guided by a learned value function. It is trained, targets CAD parts drawn with construction lines, and outputs a feature-based CAD model rather than named semantic parts; \ours{} needs no training and works on freehand sketches of everyday objects. HandMade~\cite{huang2026handmade} is closest in spirit: users label groups of VR strokes as named parts, the strokes are rendered to multi-view guidance for an image model, and the result is lifted to a textured mesh. It requires manual part labels, outputs one mesh, and lists per-part regeneration as future work; it is evaluated on custom sketches with a user study. \ours{} needs no part labels, reads the strokes as coordinates (renders optional), and produces separately editable parts; we evaluate on the public VRSketch2Shape benchmark. Interactive VR sketching systems such as SymbiosisSketch~\cite{arora2018symbiosissketch} and CASSIE~\cite{yu2021cassie} turn strokes into curves and surfaces geometrically, and a long line of work reconstructs 3D shapes from 2D sketches~\cite{igarashi1999teddy,delanoy2018sketching,zhang2021sketch2model,guillard2021sketch2mesh,gao2022sketchsampler,bandyopadhyay2024doodle,lee2025pasta}.

\paragraph{LLMs that write 3D programs.}
LLM agents have been used to set parameters of procedural generators~\cite{sun20233dgpt}, lay out scenes~\cite{hu2024scenecraft}, edit Blender materials and lighting~\cite{huang2024blenderalchemy,gu2025blendergym}, and compose objects from primitives~\cite{yamada2025l3go,zhang2025shapecraft}. LL3M~\cite{lu2025ll3m} uses several LLM agents, among them a critic that inspects renders for discrepancies with the prompt, to write modular Blender code from text and already supports follow-up edits that change only the relevant code; its evaluation reports code statistics and qualitative edits but no geometric accuracy against ground truth. Nova3D~\cite{noor2026nova3d} generates Blender source with named parts and joints from text, images or 2D sketches and reports that human reviewers found locality preserved in its edits. Treellis~\cite{chen2026treellis} has an LLM generate a tree of primitives, Boolean operations and repeats from a text prompt and exposes it for parameter edits and for node-scoped LLM edits that replace only the selected subtree, so locality holds by construction. Benchmarks such as 3DCodeBench~\cite{gao2026codebench} and 3DHarnessBench~\cite{liu2026harnessbench} evaluate frontier models writing Blender code from text, images or renders of a target object. A parallel line trains models that map text, images or point clouds to CAD or Blender code~\cite{du2024blenderllm,khan2024text2cad,li2025cadllama,rukhovich2025cadrecode,guan2025cadcoder,doris2025cadcoder,kolodiazhnyi2026cadrille,joglekar2026ortho2cad,dai2025meshcoder}, possibly with verification loops~\cite{alrashedy2025cadcodeverify,mallis2025cadassistant} or trained local editing~\cite{yuan2025caded,zhang2025flexcad}. MeshCoder~\cite{dai2025meshcoder} is the closest trained counterpart: it maps dense point clouds to part-decomposed Blender code. Except in 3DHarnessBench, which also offers full 3D access to the target object, the LLM-based systems receive no 3D input geometry, so their output is built in a frame of its own and cannot be compared with the input in 3D; MeshCoder needs dense point clouds and training. \ours{} receives a sparse, imprecise 3D sketch in a normalized metric frame and builds in that frame, which registers the output to the input and lets a geometric check compare it with the strokes and drive part-level repairs (\cref{sec:verify}), where LL3M's critic compares renders with a text prompt. The closest controlled analogues of an image-input system and of a plain LLM-to-Blender prompt are our views-only and minimal-prompt conditions (\cref{sec:ablation}). We also differ in the evaluation, a public sketch-to-shape benchmark against a trained state-of-the-art model, and in measuring edit locality mechanically, at the level of identical triangles, against a re-sketching baseline.

\paragraph{Geometry as text.}
LLaMA-Mesh~\cite{wang2024llamamesh} and MeshLLM~\cite{fang2025meshllm} fine-tune LLMs to read and write meshes serialized as text, and ShapeLLM-Omni~\cite{ye2025shapellmomni} uses learned 3D tokens. SGP-Bench~\cite{qiu2025sgpbench} probes whether LLMs understand symbolic graphics programs without rendering, SketchAgent~\cite{vinker2025sketchagent} lets an off-the-shelf model draw 2D sketches through a string-based language, and 3DrawAgent~\cite{xiao2026drawagent} has an LLM draw 3D curves from text. We use the reverse direction with an unmodified model: 3D strokes are given as coordinates and the model writes a program. Our ablation shows that this text channel carries more usable signal than rendered views of the same strokes.

\paragraph{Part-structured shapes.}
Shape programs~\cite{sharma2018csgnet,tian2019learning,jones2020shapeassembly,jones2021shapemod,jones2023shapecoder,jones2026shapelib} and hierarchical part models~\cite{mo2019structurenet} represent objects as structured, editable compositions, but are learned for fixed domains or languages. Recent part-aware generators produce part meshes from images~\cite{chen2025partgen,lin2025partcrafter,yang2025omnipart,tang2025partpacker,dong2025copart,he2026unipart}, decompose or complete existing shapes~\cite{yan2026xpart,yang2026holopart}, or segment them after the fact~\cite{liu2025partfield,yang2024sampart3d,yuan2024cadtalk}, and language-driven tools edit existing part-segmented shapes~\cite{ganeshan2024parsel,pang2026visculpt}. Large image-to-3D models~\cite{xiang2025trellis,zhao2025hunyuan3d,hunyuan2025hunyuan3d21} produce detailed but monolithic geometry. In \ours{}, parts are named and separate by construction, because the program builds them that way.

\section{Training-Free 3D-to-3D Generation}
\label{sec:method}

Given a VR sketch $\mathcal{S}=\{s_i\}_{i=1}^{N}$, where each stroke $s_i$ is a 3D polyline, and an optional short text $t$ naming the object, \ours{} produces a program $P$ whose execution yields a set of named parts $\{(n_k, M_k)\}_{k=1}^{K}$, where $n_k$ is a part name and $M_k$ a triangle mesh. No component is trained. The design principle is to stay in 3D: the sketch's geometry reaches the LLM as 3D coordinates, and the output is 3D geometry built in the sketch's own frame rather than lifted back from images. \Cref{fig:teaser} gives an overview.

\subsection{3D in: strokes as coordinates}
\label{sec:serialize}
VR sketches in public datasets carry no physical units and are expressed in arbitrary frames. We first rescale the sketch so that the longest side of its bounding box measures \SI{100}{cm}, and place the origin at the bottom center of the bounding box, with $X$ to the right, $Y$ pointing back, $Z$ up, and the object's front facing $-Y$. Each stroke is subsampled uniformly to at most 12 points (both endpoints kept), and coordinates are rounded to \SI{0.1}{cm}. The strokes are then written as a compact JSON list,
\begin{center}
\small\texttt{[\{"id":1,"points\_cm":[[-21.4,3.0,0.0],...]\},...]}
\end{center}
together with the bounding-box size in centimeters. Strokes keep the order in which they were drawn. A test sketch has a median of 80 strokes and becomes a median of 10.8k characters of text.

By default, as in the VR application the prompt was written for, the request also carries five orthographic line drawings of the sketch (front, left, back, right and top; $512\times512$, blue lines on white, hidden lines visible), all rendered at a common scale. The ablation in \cref{sec:ablation} shows that removing them does not change the F-score significantly, whereas replacing the coordinates by them costs 7.8 F-score points; unless marked ``coordinates only'' (no views; the category name is still given), all results use both.

\subsection{3D out: a part program in the sketch's frame}
\label{sec:synthesis}
An LLM receives a fixed system prompt and a user message, and replies with a complete Blender~3.6 Python script. The system prompt (\cref{app:prompts}) sets the role (``a 3D artist who turns a user's rough mid-air 3D sketch into a clean, realistic model by writing a Blender script'') and the rules for the program:
\begin{itemize}
  \item build in the same frame as the serialized strokes, in meters (the strokes are in centimeters);
  \item build the object as \emph{separate} mesh objects, one per meaningful editable part, each with a short unique PascalCase name, using 3 to 12 top-level parts; small details are parented to the part they belong to; parts are never joined;
  \item assemble the parts into one object without floating parts or gaps, and do not extend beyond the sketch's outline by more than about \SI{2}{cm} unless the strokes show it;
  \item use clean geometry (bevels, subdivision, lathe) and constant-valued materials, and add no cameras or lights.
\end{itemize}
The program works in the same frame as the serialized strokes, so inverting the canonicalization of \cref{sec:serialize} places the result exactly where, and at the size, the user drew it; no registration step is needed (\cref{sec:registration}). The user message states that the user drew a rough 3D sketch in mixed reality, gives the text $t$ (e.g., ``a chair''), the sketch size, the serialized strokes, and a shape-fidelity instruction. We use the ``balanced'' level: \emph{follow the strokes' overall size, proportions and where each part sits closely, but clean up and refine the shapes so the object looks well made.}

The prompt was written for an interactive VR modeling application before this study, and we use it unchanged. It contains rules unrelated to the benchmark categories (\eg, how to build animated characters). We did not tune it on the benchmark.

\subsection{Execution}
\label{sec:execute}
The script is executed in a headless Blender~3.6. If it raises an error, or produces fewer than two top-level parts, the error message and the script are sent back to the LLM once, with a request for the full fixed script. Each top-level mesh object (with its parented details merged into it) becomes one part $(n_k, M_k)$, with a normalized, unique name. Execution is deterministic: building the same script twice yields the same set of triangles (in a different order).

\subsection{Part-level editing}
\label{sec:edit}
Because the output is a program with named parts, an edit can address one part by name. Given $P$, a part name $n$ and a text instruction (\eg, ``Make this part about 40\% bigger in every direction, and keep it attached where it is attached now''), the LLM is asked to return the full script with only the requested change applied, keeping every other part byte-for-byte identical in the code, with the same object names (\cref{app:prompts}). The new script is executed as above. No mechanism enforces locality: the LLM rewrites the whole script, and we \emph{measure} whether untouched parts survive (\cref{sec:editing}).

\subsection{Checking the result against the sketch}
\label{sec:verify}
Because the parts are built in the sketch's frame, the result can be compared with the strokes directly, without ground truth and without registration. We resample the strokes every \SI{0.5}{cm} and sample the surface of every part. A stroke point is \emph{covered} if it lies within \SI{2}{cm} (2\% of the sketch size) of the surface, and a stroke is \emph{missed} if most of its points are not covered. A surface point is \emph{supported} if it lies within \SI{3}{cm} of a stroke, or if strokes in its own plane surround it: within \SI{60}{cm}, stroke points that are less than \SI{2}{cm} from its tangent plane, on strokes running parallel to that plane, lie in all four quadrants around it. The second condition accepts the middle of a door or table top whose outline is drawn and rejects an interior shelf, which is crossed only by vertical corner strokes. For each part we compute the share of its surface that is supported and the length of the strokes for which it is the nearest part.

If a stroke is missed, or a part covering at least 3\% of the surface has little support, the evidence is sent through the editing mechanism of \cref{sec:edit}, without naming a target part: the missed strokes with their coordinates and nearest part, and the support of every part (an example is in \cref{app:prompts}). The LLM decides which parts to move, reshape or delete; all other parts must stay unchanged. The repaired model is kept only if it raises the check score $c + (A_s - A_u)/A_0$ by more than 0.01 and lowers $c$ by at most 0.01, where $c$ is the share of covered stroke points, $A_s$ and $A_u$ are the supported and unsupported surface areas, and $A_0$ is the area before the repair. Using areas rather than shares means that deleting a supported part lowers the score. We run at most two rounds; all thresholds are listed in \cref{app:checkdesign}. Two simpler acceptance scores, including the mean distance between strokes and surface in both directions, work as well (\cref{sec:verify_results}); what matters is that output and strokes can be compared at all. Ground truth is used only to evaluate the result.

\section{Experimental Setup}
\label{sec:setup}

\paragraph{Data.}
We use the hand-drawn test set of VRSketch2Shape~\cite{chen2026vrsketch2shape}: 400 VR sketches of 400 different shapes, 100 each of airplanes, cabinets, chairs and tables, drawn by participants in VR. For each shape we use the detailed sketch provided with the dataset. Ground-truth surfaces are extracted from the dataset's $64^3$ signed distance fields (clipped to $\pm0.2$) by marching cubes at level $0.02$, following the official evaluation code.

\paragraph{Baseline.}
VRSketch2Shape~\cite{chen2026vrsketch2shape} is the state of the art on this benchmark: a sketch-conditioned latent diffusion model built on SDFusion~\cite{cheng2023sdfusion} and trained on paired VR sketches and shapes; it outputs a $64^3$ SDF. We run the official multi-class checkpoint with the official inference settings (DDIM, 100 steps, $\eta=0$, initial noise seeded with $10$ plus the sample index) and extract its surface at level $0.005$ as the official code does. The official inference code requires CUDA-only components; we ported it to Apple MPS by replacing the stroke simplification library with a line-by-line port (verified point by point against the original) and moving the sampler to the model's device. Over all 400 sketches our run reproduces the published F-score exactly (69.8). The mean Chamfer distance differs from the published value (4.93 vs.\ 14.8); this mean is dominated by a few outliers (our median is 2.67, maximum 229), so we also report medians.

\paragraph{Metrics.}
We follow the official protocol. 4096 points are sampled on the predicted and ground-truth surfaces, each point cloud is normalized independently to the $[-1,1]$ bounding box, and we report the Chamfer distance (CD, mean squared nearest-neighbor distance summed over both directions, $\times10^3$) and the F-score at $\tau=0.04$~\cite{tatarchenko2019single}. Because the evaluation normalizes each shape to its own bounding box, it is insensitive to the absolute scale and placement of the output. We additionally report normal consistency (NC)~\cite{mescheder2019occupancy} on 30k points, and CD/F-score after aligning the prediction to the ground truth with a symmetric ICP restricted to translation and uniform scale (scale clipped to $[0.8, 1.25]$). Differences between methods are tested with paired Wilcoxon signed-rank tests over sketches and computed from unrounded values; per-category $p$-values are exploratory and not corrected for multiple comparisons.

\paragraph{Implementation.}
The LLM is GPT-6 Astra~\cite{openai2026gpt6astra} (model identifier \texttt{gpt-6-astra}), queried through the OpenAI Responses API with default settings between October~28 and November~15, 2026. The text $t$ is the category name (``a chair''); \cref{sec:ablation} also reports results without it, since the baseline does not receive the category. Programs are executed with Blender~3.6 on a 2-vCPU cloud server. Unless stated otherwise, each sketch is run once.

\section{Results}
\label{sec:results}

\subsection{Comparison with other methods}
\label{sec:main}

\begin{table}[t]
  \centering
  \small
  \setlength{\tabcolsep}{2.7pt}
  \begin{tabular}{@{}llccc@{}}
    \toprule
    Method & Input & $n$ & F $\uparrow$ & \ours{} \\
    \midrule
    \multicolumn{5}{@{}l}{\textit{Trained on paired sketches and shapes}} \\
    Luo \etal~\cite{luo2023vrsketch}$^\dagger$ & point cloud & 400 & 48.8 & 71.3 \\
    LAS-Diffusion~\cite{zheng2023lasdiffusion}$^\dagger$ & 5 views & 400 & 40.2 & 71.3 \\
    VRSketch2Shape~\cite{chen2026vrsketch2shape} & stroke sequence & 400 & 69.8 & 71.3 \\
    \midrule
    \multicolumn{5}{@{}l}{\textit{Training-free}} \\
    2D detour & 1 view & 100 & 43.3 & 71.3$^\ddagger$ \\
    Views only & 5 views & 40 & 62.8 & 70.6$^\ddagger$ \\
    Minimal prompt$^\S$ & coordinates & 40 & 65.9 & 70.6$^\ddagger$ \\
    LL3M-style agent$^\P$ & coord.\ + views & 40 & 72.2 & 70.6$^\ddagger$ \\
    DeepSeek-V4-Pro & coordinates & 60 & 40.6 & 71.0$^\ddagger$ \\
    Claude Fable 5.2 & coordinates & 60 & 68.3 & 71.0$^\ddagger$ \\
    Gemini 4 Argon & coordinates & 60 & 68.2 & 71.0$^\ddagger$ \\
    \ours{} & coord.\ + views & 400 & 71.3 & -- \\
    \bottomrule
  \end{tabular}
  \caption{\textbf{All comparisons at a glance} (F-score, official protocol). $n$: number of test sketches compared (fixed subsets below 400). ``Views'': renderings of the strokes. Last column: \ours{} on the same sketches; $^\ddagger$with coordinates only and the same prompt. $^\dagger$Reported by~\cite{chen2026vrsketch2shape} for models retrained on the same data; our reproduction of their own model gives their 69.8 exactly. $^\S$GPT-6 Astra told only to write a Blender script, its ground plane removed. $^\P$Our re-implementation of LL3M's~\cite{lu2025ll3m} agent loop with the same LLM (\cref{s:further}). Details: \cref{sec:main} (trained), \cref{sec:ablation} (views), \cref{sec:openllm} (other LLMs), \cref{sec:detour} (2D detour); outputs of the main methods on the same sketches in \cref{app:routes}.}
  \label{tab:routes}
\end{table}

\begin{table}[t]
  \centering
  \small
  \setlength{\tabcolsep}{3.2pt}
  \resizebox{\linewidth}{!}{\begin{tabular}{@{}lccccccc@{}}
    \toprule
    & \multicolumn{3}{c}{F-score $\uparrow$} & \multicolumn{2}{c}{CD mean $\downarrow$} & \multicolumn{2}{c}{CD median $\downarrow$} \\
    \cmidrule(lr){2-4}\cmidrule(lr){5-6}\cmidrule(l){7-8}
    Category & Trained & Ours & wins & Trained & Ours & Trained & Ours \\
    \midrule
    Airplane & 86.6 & \textbf{92.3}$^{***}$ & 74 & 1.81 & \textbf{1.28} & 1.49 & \textbf{1.07}$^{***}$ \\
    Cabinet  & \textbf{63.1}$^{***}$ & 51.4 & 30 & \textbf{7.85} & 11.17 & \textbf{3.83}$^{***}$ & 7.12 \\
    Chair    & 68.8 & \textbf{72.3}$^{**}$ & 63 & \textbf{3.29} & 3.84 & 2.88 & \textbf{2.69} \\
    Table    & 60.6 & \textbf{68.9}$^{***}$ & 73 & 6.75 & \textbf{4.28} & 3.84 & \textbf{3.27}$^{**}$ \\
    \midrule
    All (400) & 69.8 & \textbf{71.3}$^{**}$ & 240 & \textbf{4.93} & 5.14 & \textbf{2.67} & 2.83 \\
    \bottomrule
  \end{tabular}}
  \caption{\textbf{\ours{} vs.\ the trained VRSketch2Shape~\cite{chen2026vrsketch2shape}} on all 400 hand-drawn test sketches (100 per category), official protocol. CD is $\times10^3$. ``wins'': sketches (of 100, or of 400) on which \ours{} has the higher F-score. Bold: the better value. Stars: paired Wilcoxon signed-rank test over sketches (for CD one test, marked in the median column), $^{**}p<0.01$, $^{***}p<0.001$; unmarked differences are not significant at $p<0.05$.}
  \label{tab:main}
\end{table}

\Cref{tab:routes} collects all comparisons. Among trained methods, VRSketch2Shape is far ahead of the two earlier ones that~\cite{chen2026vrsketch2shape} retrained on the same data and evaluated with the same protocol: Luo \etal~\cite{luo2023vrsketch}, which reads the strokes as a point cloud, and LAS-Diffusion~\cite{zheng2023lasdiffusion}, which reads rendered views. The training-free routes are compared on fixed subsets in \cref{sec:ablation,sec:openllm,sec:detour}.

\begin{figure*}[t]
  \centering
  \includegraphics[width=\linewidth,height=0.78\textheight,keepaspectratio]{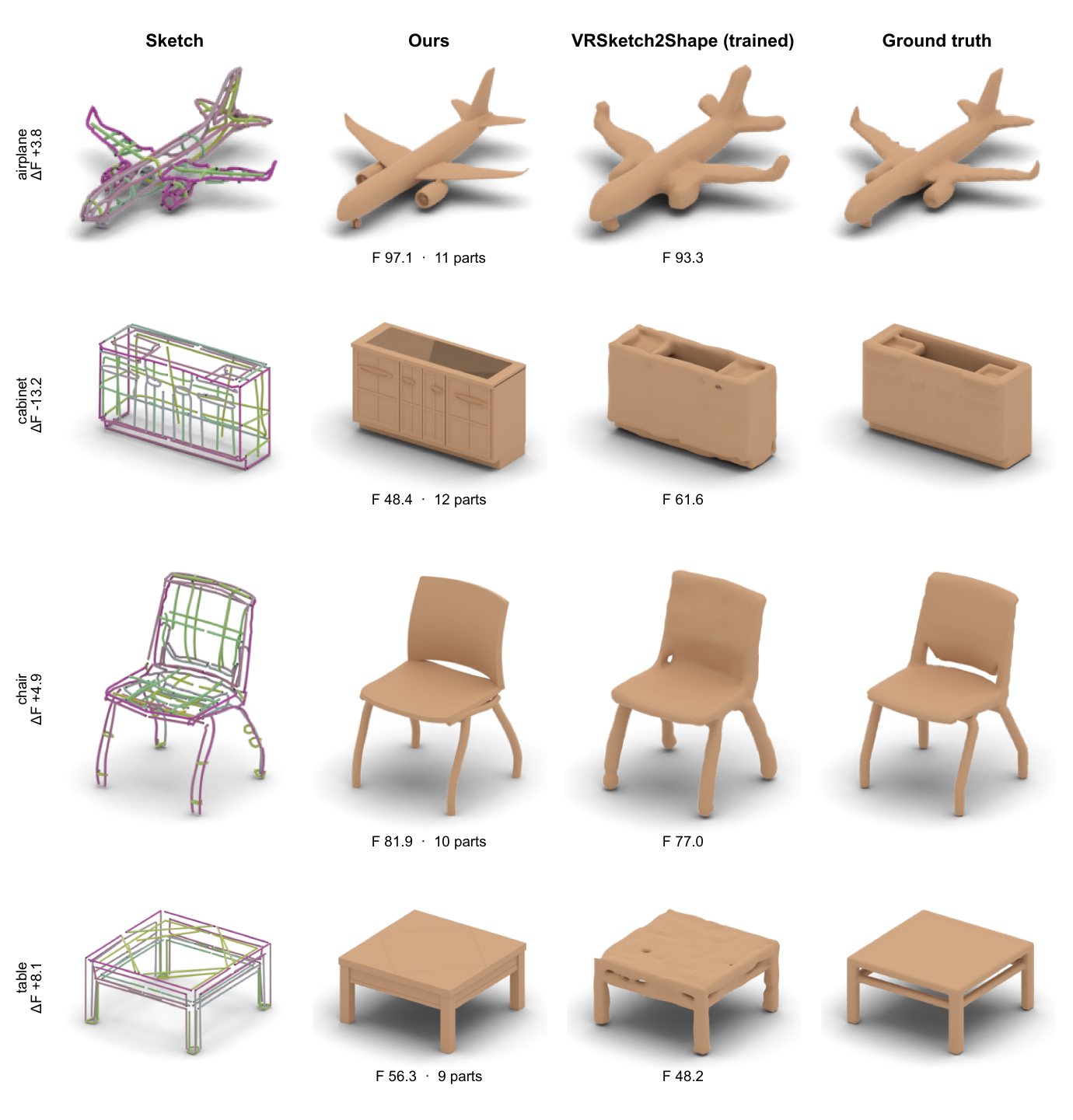}
  \caption{\textbf{Typical results.} For each category, the sketch at the median F-score difference between \ours{} and the trained model ($\Delta$F, left), so the examples are not selected for appearance. Our programs are rendered in the same color as the other shapes, and strokes are colored by drawing order, from purple (first) to yellow (last); each shape is normalized to its own bounding box as in the evaluation. The best and worst cases per category are in \cref{app:qual}.}
  \label{fig:qual}
\end{figure*}

\Cref{tab:main} compares \ours{} with the trained model on the full test set. Without any training, \ours{} obtains a higher F-score overall (71.3 vs.\ 69.8; higher on 240 of 400 sketches; $p=0.005$). The effect is small: the median per-sketch difference is 2.7 points (95\% bootstrap interval of this median $[1.6, 4.0]$) and the rank-biserial correlation 0.16. \ours{} is significantly better on airplanes, chairs and tables, and significantly worse on cabinets (\cref{sec:cabinets}), on which it loses 11.6 points on average and which therefore make the mean difference (1.5, interval $[-0.1, 3.1]$) smaller than the median. The paired Wilcoxon test compares the per-sketch differences by rank and is not a test of the mean, whose interval includes zero. The Chamfer distance shows no significant overall difference (median 2.83 vs.\ 2.67, $p=0.71$); its mean is dominated by a few cabinets. Every one of the 400 sketches produced a valid part program; 6 of them needed the single repair round after a Blender error. The programs have a median of 10 named parts (range 3--12). \Cref{fig:qual} shows examples drawn at the best, median and worst F-score difference per category.

\subsection{Beyond the official metrics}
\label{sec:more_metrics}

\begin{table}[t]
  \centering
  \small
  \setlength{\tabcolsep}{4pt}
  \resizebox{\linewidth}{!}{\begin{tabular}{@{}lccccc@{}}
    \toprule
    & F aligned $\uparrow$ & CD aligned $\downarrow$ & F exposed $\uparrow$ & NC $\uparrow$ & NC exposed $\uparrow$ \\
    \midrule
    Trained & \textbf{75.3} & \textbf{4.07}$^{*}$ & 69.7 & \textbf{88.6}$^{***}$ & \textbf{88.6}$^{***}$ \\
    \ours{} & 73.7 & 4.48 & \textbf{73.0}$^{***}$ & 84.2 & 85.2 \\
    \bottomrule
  \end{tabular}}
  \caption{\textbf{Additional metrics} on all 400 sketches. ``Aligned'': after a symmetric ICP restricted to translation and uniform scale. NC: normal consistency~\cite{mescheder2019occupancy}. ``Exposed'' samples points only on the exterior of each shape (surface points within 2 voxels of the outside of the filled solid at $256^3$), applied identically to both methods; we defined it post hoc, after inspecting the outputs, and treat it as a diagnostic. The aligned F-scores do not differ significantly ($p=0.43$). Stars as in \cref{tab:main}, with $^{*}p<0.05$.}
  \label{tab:more}
\end{table}

The official protocol normalizes each shape to its own bounding box, so a small protrusion or misplacement shifts the whole shape. After aligning each prediction to the ground truth by translation and uniform scale (\cref{tab:more}), the trained model gains more than \ours{} (69.8$\to$75.3 vs.\ 71.3$\to$73.7), and the F-scores no longer differ significantly ($p=0.43$), while the trained model has a lower aligned Chamfer distance (mean 4.07 vs.\ 4.48, $p=0.03$), owing to cabinets alone (without them \ours{} is lower, 2.80 vs.\ 3.27, $p=0.03$). By category, \ours{} remains significantly better on airplanes (93.6 vs.\ 92.1) and tables (71.8 vs.\ 68.3), ties on chairs, and is worse on cabinets. \ours{} has lower normal consistency (84.2 vs.\ 88.6, $p<10^{-20}$), except on airplanes, where it is better on the exposed surface (88.9 vs.\ 86.9). The ground truth is a smooth $64^3$ SDF; the trained model's surfaces are similarly smooth, whereas our programs add crisp edges, bevels and small components whose normals depart from it. Restricting the evaluation to the exterior of each shape widens the F-score gap in favor of \ours{} (73.0 vs.\ 69.7, higher on 274 of 400 sketches, $p<10^{-6}$), consistent with interior structure costing it points (\cref{sec:cabinets}). We therefore summarize the geometric comparison as \emph{better on the official F-score and not significantly different in F-score after alignment}.

\subsection{Categories and shapes the trained model has not seen}
\label{sec:ood}
The benchmark's four categories are the ones the trained model was trained on. To test others, we synthesized line-drawing sketches (creases, boundaries and silhouettes, with a hand-like wobble) from 36 ModelNet10 test shapes~\cite{wu2015shapenets}: six each of bed, monitor, toilet and bathtub, which resemble none of the training categories, and of chair and table as a control for the synthetic drawing style (\cref{s:ood}). The reference surfaces had to be rebuilt; we built them like the benchmark's, as offset surfaces of a $64^3$ distance field, and also scored against the raw meshes.

On the unseen categories, \ours{} scores higher than the trained model on 20 of 24 sketches against the rebuilt references (50.4 vs.\ 38.2, $p<0.001$) and on 23 of 24 against the raw meshes (65.5 vs.\ 36.8, $p<0.001$), and also at the looser thresholds F@0.06 and F@0.1 and in Chamfer distance (\cref{tab:ood}). The trained model returns shapes that fill the extent of the strokes without their structure, while \ours{} returns recognizable beds, monitors, toilets and bathtubs (\cref{fig:ood}). On the control categories, the ranking depends on the reference: the trained model leads against the rebuilt references (66.2 vs.\ 53.1), \ours{} against the raw meshes (77.7 vs.\ 46.0). The rebuilt references favor the trained model, which was trained to reproduce such offset surfaces: a perfect reconstruction, the ModelNet10 mesh itself, scores only 53.1 against them on the control categories, the same as \ours{} and below the trained model. Nor is the control gap due to parts our programs miss: 3.3\% of the reference surface lies more than 5\% of the object size from our outputs, against 5.7\% for the trained model. From the control to the unseen categories, the trained model falls from 66.2 to 38.2 and \ours{} from 53.1 to 50.4 (drops differ, $p<0.001$); against the raw meshes, both fall by about ten points, and the drops differ only at F@0.1 (\cref{tab:ooddrops}). Thirty-six synthetic sketches do not measure performance on new categories drawn by people; they show that the method is not limited to the benchmark's categories, while the trained model degrades outside them.

The trained model's drop could by itself make \ours{} look good on the unseen categories. The 2D detour (\cref{sec:detour}) is built from general-purpose image and image-to-3D models trained on no particular category; run unchanged on the same sketches, it reaches 27.4 on the unseen categories against the rebuilt references and 31.2 against the raw meshes, and \ours{} is better on 23 and 24 of the 24 sketches ($p<10^{-5}$; \cref{tab:ood}). The detour also drops from the control categories (36.0 and 46.5; not significant), so part of every method's drop comes from these categories and their synthetic sketches, but neither baseline matches \ours{} on them.

The six categories above are rigid objects. On 20 synthetic sketches of people and plants from ModelNet40, irregular shapes far from the benchmark's, all our programs built, as people whose head, torso, arms and legs follow the sketched pose and plants with pots, stems and leaves modeled one by one (\cref{fig:organic_all}). Against the raw meshes, \ours{} scores 68.9, the trained model 36.0 and the 2D detour 33.1, and \ours{} is better than each on 19 of the 20 sketches ($p<10^{-5}$; \cref{tab:organic}). Our prompt, not changed for this test, has general rules for people but none for plants, which show the same advantage (65.9 vs.\ 30.9). Against the rebuilt references, \ours{} still beats the detour (52.0 vs.\ 33.0, $p<0.001$), but its lead over the trained model is not significant (52.0 vs.\ 45.0, $p=0.10$): the offset surface thickens thin limbs, stems and leaves, so that the meshes themselves score only 68.4 against it. The detour's image model draws plausible people and plants, but often not the sketched ones (\cref{fig:organic}).

\subsection{3D in: coordinates versus views}
\label{sec:ablation}

\begin{table}[t]
  \centering
  \small
  \setlength{\tabcolsep}{4pt}
  \resizebox{\linewidth}{!}{\begin{tabular}{@{}llccc@{}}
    \toprule
    Subset & Input to the LLM & F $\uparrow$ & CD mean $\downarrow$ & vs.\ trained \\
    \midrule
    \multirow{5}{*}{100} & coordinates + views + label (default) & 71.6 & 4.13 & $p=0.50$ \\
    & coordinates + label & 71.3 & 4.37 & $p=0.76$ \\
    & coordinates + views, \emph{no label} & 72.2 & 4.29 & $p=0.19$ \\
    & \textit{2D detour (image-to-3D)} & \textit{43.3} & \textit{17.56} & $p<10^{-14}$ \\
    & \textit{trained model} & \textit{71.1} & \textit{3.95} & -- \\
    \midrule
    \multirow{5}{*}{40} & coordinates + views + label & 69.1 & 4.20 & $p=0.93$ \\
    & coordinates + label (3D only) & 70.6 & 4.50 & $p=0.72$ \\
    & views + label (2D only) & 62.8 & 6.72 & $p=0.013$ \\
    & coordinates + label, minimal prompt$^\S$ & 65.9 & 19.54 & $p=0.31$ \\
    & \textit{trained model} & \textit{69.0} & \textit{4.81} & -- \\
    \bottomrule
  \end{tabular}}
  \caption{\textbf{Input ablation} on fixed random subsets (25 or 10 sketches per category). ``vs.\ trained'': paired Wilcoxon test on F-score against the trained model on the same sketches. Views alone are significantly worse than both the trained model and the full input ($p<0.001$); dropping the views or the category label makes no significant difference ($p=0.24$, $p=0.54$). The 2D detour renders the sketch, generates an image from it and lifts the image to 3D without an LLM (\cref{sec:detour}). $^\S$Same LLM and data, with only the instruction to write a Blender script; the ground plane it adds to every scene is removed (8.1 with it).}
  \label{tab:ablation}
\end{table}

\Cref{tab:ablation} varies what the LLM receives, with the same model and prompt. This isolates the 2D detour taken by training-free pipelines that render the sketch for an image model: the views are orthographic renderings of exactly the same strokes. Removing the rendered views does not change the F-score significantly (71.3 vs.\ 71.6, $p=0.24$). Replacing the coordinates by the views costs 7.8 F-score points (70.6 vs.\ 62.8 on the same 40 sketches; coordinates better on 31 of 40, $p<10^{-4}$) and makes \ours{} significantly worse than the trained model. The 3D coordinates therefore carry almost all of the signal; rendering the sketch to 2D throws it away. Our prompt adds to this but is not its source: given the same coordinates and only the instruction to write a Blender~3.6 script (\cref{app:prompts}) that builds the object, the same LLM reaches 65.9 on these 40 sketches, above the views with our full prompt ($p=0.02$) and below the coordinates with our full prompt (70.6, ours better on 27 of 40, $p=0.004$), mostly on chairs (51.5 vs.\ 65.5). This counts only the object: with that minimal prompt the LLM adds a ground plane to every scene (8.1 with it), and it returns a median of 36.5 separate objects instead of 10 named parts. Nor does the result hinge on our pipeline. LL3M~\cite{lu2025ll3m} cannot be run (only its client is public, and its coding model has been retired), so we re-implemented its loop with the same LLM and coordinates (\cref{s:further}): a planning agent decomposes the object into components with dimensions and placement, a coding agent writes the script, build errors go back to it, a critic compares renderings of the result with renderings of the sketch from four views, and the script is revised once. On the same 40 sketches it reaches 72.2, not significantly different from our 70.6 ($p=0.18$, each better on 20) and from the trained model ($p=0.24$); its critique round raises it from 71.5 to 72.2 ($p=0.03$). It makes 3.6 calls and costs \$0.83 per sketch, 2.9$\times$ as much as \ours{} with the same input, and returns a median of 41 objects instead of 10 parts. Two pipelines of different design thus reach a comparable level from the same coordinates: what matters is giving the LLM the strokes as coordinates in the frame in which it builds. Because our default prompt names the category and the trained model does not receive it, we also ran \ours{} with no text at all (the prompt then states that the user said nothing and asks the LLM to infer the object from the sketch). This does not lower the score (72.2 vs.\ 71.6, $p=0.54$), so the comparison in \cref{tab:main} is not driven by the label.

\paragraph{Missing strokes.}
To test incomplete sketches, we dropped 30\% of the strokes of each of the 40 sketches at random (seed fixed per sketch, drawing order kept; a median of 56 of 80 strokes remain) and gave both methods the same reduced sketch; for \ours{} the frame is recomputed from the remaining strokes, as it would be if the user had not drawn the others. With coordinates only, our F-score falls from 70.6 to 67.3 ($p=0.047$; median change $-1.6$) and the trained model's from 69.0 to 67.3 ($p=0.055$); the drops do not differ significantly ($p=0.48$), and the two still do not differ significantly (67.3 vs.\ 67.3, $p=0.94$). Airplanes lose the most for both (ours 89.5 to 80.4, trained 89.5 to 82.1), and cabinets do not drop (\cref{tab:dropout}).

\begin{table}[t]
  \centering
  \footnotesize
  \setlength{\tabcolsep}{3pt}
  \begin{tabular}{@{}lccccc@{}}
    \toprule
    & Air. & Cab. & Chair & Table & All \\
    \midrule
    \ours{}, all strokes & 89.5 & 61.5 & 65.5 & 65.9 & 70.6 \\
    \ours{}, 30\% dropped & 80.4 & 61.9 & 64.0 & 63.0 & 67.3 \\
    Trained, all strokes & 89.5 & 61.7 & 66.8 & 57.9 & 69.0 \\
    Trained, 30\% dropped & 82.1 & 65.6 & 64.6 & 56.9 & 67.3 \\
    \bottomrule
  \end{tabular}
  \caption{\textbf{Missing strokes} (F-score) on the 40 sketches of \cref{tab:ablation} (10 per category), with all strokes and with the same 30\% of the strokes dropped for both methods; \ours{} with coordinates and label, no views.}
  \label{tab:dropout}
\end{table}

\paragraph{How the coordinates are written.}
\label{sec:serialization}
The strokes reach the LLM as text: per stroke, up to 12 points subsampled along it, in drawing order, in centimeters with one decimal. To see whether the results depend on this serialization, we reran the same 40 sketches with the coordinates-plus-label input three times, changing one choice each time: the strokes in a random order (fixed per sketch), at most 6 points per stroke, or coordinates rounded to whole centimeters (\cref{tab:serialization}). The F-score stays within 1.6 points of the default (70.3, 69.6 and 69.0 vs.\ 70.6); only rounding is nominally lower ($p=0.04$, not significant after correcting for three tests), and no variant differs significantly from the trained model (69.0; $p\geq0.76$). All 120 programs built. The method does not depend on a particular serialization; fewer points or coarser coordinates shorten the input by about 30\% (medians over these 40 sketches: 7.3k and 7.5k instead of 10.4k characters) at a small cost.

\begin{table}[t]
  \centering
  \footnotesize
  \setlength{\tabcolsep}{3pt}
  \resizebox{\linewidth}{!}{\begin{tabular}{@{}lcccccc@{}}
    \toprule
    & Air. & Cab. & Chair & Table & All & vs.\ default \\
    \midrule
    Default & 89.5 & 61.5 & 65.5 & 65.9 & 70.6 & -- \\
    Strokes shuffled & 91.4 & 61.8 & 63.8 & 64.0 & 70.3 & $p=0.67$ \\
    At most 6 points & 89.8 & 60.3 & 60.6 & 67.8 & 69.6 & $p=0.28$ \\
    Rounded to 1 cm & 89.9 & 57.1 & 65.3 & 63.8 & 69.0 & $p=0.04$ \\
    \textit{Trained model} & \textit{89.5} & \textit{61.7} & \textit{66.8} & \textit{57.9} & \textit{69.0} & \\
    \bottomrule
  \end{tabular}}
  \caption{\textbf{Coordinate serialization} (F-score) on the 40 sketches of \cref{tab:ablation}, coordinates and label, the same prompt. Default: at most 12 points per stroke, 0.1 cm resolution, strokes in drawing order; one choice changed per row; paired Wilcoxon test against the default.}
  \label{tab:serialization}
\end{table}

\paragraph{Rotated sketches and missing components.}
To test whether the output follows the sketch where it is drawn, we rotated the 40 sketches of \cref{tab:serialization} about the vertical axis, together with their references, and, separately, removed every stroke of one component, such as a wing, a leg or a side panel, chosen at random (12\% of the strokes on average), keeping the full reference (\cref{s:further}). Rotated by \ang{45}, our F-score with the default input (coordinates and views) falls from 69.1 to 64.2 ($p=0.02$); by \ang{90}, which only exchanges axes, it does not fall (70.7). The trained model generates in its canonical orientation and falls to 43.7 and 47.2 ($p<10^{-7}$ for both), so \ours{} is better on 36 and 35 of the 40 rotated sketches. Rotated back, our output agrees with the output for the unrotated sketch at an F-score of 75.0 (\ang{45}) and 79.8 (\ang{90}), close to the 81.7 between two runs on the same unrotated sketch; the trained model's agrees at 44.4 and 53.8. A missing component costs us 2.2 points ($p=0.52$) and the trained model 3.9 ($p=0.005$); on these sketches the two do not differ significantly ($p=0.53$).

\paragraph{Other LLMs.}
\label{sec:openllm}
To see whether the approach is specific to GPT-6 Astra, we ran two other proprietary frontier models, Claude Fable 5.2~\cite{anthropic2026fable} and Gemini 4 Argon~\cite{google2026gemini4}, and the strongest open-weight model available to us, DeepSeek-V4-Pro~\cite{deepseek2026v4pro}, with the same prompt and the coordinates-plus-label input, through the same builder and scoring, on 60 of the 100 sketches (15 per category, in a fixed interleaved order; details in \cref{app:openllm}). On the same sketches, Claude Fable 5.2 does not differ significantly from GPT-6 Astra (68.3 vs.\ 71.0, $p=0.87$; better on 30 of 60); its lower mean comes from four replies, two airplanes and two chairs, that used up the 32k-token output limit without returning a program and count as zero, and on the 56 others it reaches 73.1 (GPT-6 Astra 71.3). Gemini 4 Argon is slightly but significantly below GPT-6 Astra (68.2 vs.\ 71.0, $p=0.01$; better on 21 of 60), mostly on chairs and airplanes (59.4 vs.\ 65.1 and 83.2 vs.\ 87.3), and one of its scripts, for a cabinet, still failed to build after the fix round. Neither differs significantly from the trained model (70.7 on the same sketches, $p=0.96$ and $p=0.34$), and their programs lie as close to the strokes as GPT-6 Astra's (median distance of a stroke point to the surface 0.56\% and 0.60\% of the sketch size, against 0.59\%). DeepSeek-V4-Pro reaches 40.6, against 71.0 for GPT-6 Astra on the same sketches (higher for GPT-6 Astra on 58 of 60, $p<10^{-10}$) and 70.7 for the trained model. Seven of its replies, five of them airplanes, used up the output limit on reasoning without returning a program; on the other 53 sketches it reaches 45.9. Its programs have as many named parts as ours (median 9 vs.\ 10) but follow the strokes far less closely: 64\% of stroke points lie within 2\% of the sketch size from its surface, against 96\% for GPT-6 Astra. With frontier models from three vendors the approach thus reaches the level of the trained model; with the open-weight model we tried it does not.

\paragraph{A complete 2D detour.}
\label{sec:detour}
The ablation keeps the LLM and changes only its input. To test a representative 2D detour, we built one from public models (details in \cref{app:detour}). The strokes are drawn as black lines from a three-quarter view (\ang{35} azimuth, \ang{20} elevation), as a user would see them; a scribble ControlNet~\cite{zhang2023controlnet} on Stable Diffusion~1.5~\cite{rombach2022ldm} turns the drawing into a product photo of the named category; after background removal, Hunyuan3D-2~\cite{zhao2025hunyuan3d} lifts the photo to a mesh. Like \ours{}, the detour receives the category name, and its mesh is scored with the same protocol as all other outputs. Image-to-3D models generate in a canonical frame of their own, so for each sketch we rotate the mesh about the vertical axis in \ang{15} steps and keep the best F-score; this oracle favors the baseline (one rotation for all sketches gives 40.8 instead of 43.3). On the 100-sketch subset of \cref{tab:ablation}, the detour reaches 43.3, against 71.3 for \ours{} with coordinates only (71.6 with views) and 71.1 for the trained model. Coordinates alone are better on 94 of 100 sketches ($p<10^{-16}$) and in every category, and the mean Chamfer distance of the detour is four times ours (17.6 vs.\ 4.1). HandMade~\cite{huang2026handmade} takes this route with stronger components, a commercial multi-view image generator and a multi-view image-to-3D model, and with part labels from the user, but reports no results on this benchmark; our detour stands for the route, not for that system, and a stronger image generator or image-to-3D model might narrow the gap. A trained model that reads rendered views lands at a similar level (LAS-Diffusion, 40.2; \cref{tab:routes}). \Cref{fig:detour} shows the typical failure. A VR sketch is a wireframe: hidden edges stay visible, and many sketchers hatch faces with diagonal strokes. In a single drawing nothing distinguishes these lines from physical structure, so the image model draws them as wires, glass panes and lattices, and the image-to-3D model builds those. On airplanes, the image model sometimes replaces the drawn design by a generic airliner or jet. In 3D, a hatching stroke lies in the plane of a face and a hidden edge lies behind it, which one view cannot show.

\paragraph{Run-to-run variation.}
LLM sampling makes \ours{} stochastic. We generated the same 100 sketches a second time (one generation failed with an API-side server error and is excluded). On the remaining 99 sketches, the mean F-score is stable (71.3 vs.\ 71.5, $p=0.58$; per-sketch correlation 0.96), and the second run leads to the same per-category conclusions against the trained model. Individual sketches vary more: the median absolute change in F-score between runs is 1.5 points, the mean 3.2, and 9\% of sketches change by more than 10 points.

\paragraph{Cost.}
A sketch costs \$0.31 on average at list prices and takes a median of \SI{58}{s} end to end, including program execution. The serialized strokes have a median length of 10.8k characters (80 strokes).

\subsection{Shape fidelity}
\label{sec:fidelity}

\begin{table}[t]
  \centering
  \footnotesize
  \setlength{\tabcolsep}{2.6pt}
  \begin{tabular}{@{}lcccccccc@{}}
    \toprule
    & \multicolumn{5}{c}{F $\uparrow$} & CD $\downarrow$ & F al.\ $\uparrow$ & NC $\uparrow$ \\
    \cmidrule(lr){2-6}
    Method & Air. & Cab. & Chair & Table & All & med. & all & all \\
    \midrule
    Balanced$^\ast$ & 89.0 & 56.5 & 66.3 & 72.4 & 71.1 & 3.13 & 74.3 & 85.4 \\
    Strict & \textbf{91.7} & 64.8 & \textbf{73.5} & \textbf{73.2} & \textbf{75.8} & \textbf{2.41} & \textbf{78.4} & 87.0 \\
    \textit{Trained} & \textit{84.4} & \textit{\textbf{69.3}} & \textit{65.9} & \textit{63.1} & \textit{70.7} & \textit{2.56} & \textit{77.7} & \textit{\textbf{89.3}} \\
    \bottomrule
  \end{tabular}
  \caption{\textbf{Shape-fidelity instruction} on 80 sketches (the first 20 per category, in identifier order, of the 100 in \cref{tab:ablation}). Balanced and Strict are \ours{} with one different sentence in the user message (\cref{app:prompts}); $^\ast$the default used in all other experiments. Trained: VRSketch2Shape. F al.: F-score after translation and uniform-scale alignment (\cref{sec:more_metrics}); NC: normal consistency. Strict versus balanced: higher F on 62 of 80 sketches ($p<10^{-7}$); strict versus the trained model: $p=0.001$ (F), $p=0.45$ (F al.), $p=0.003$ (NC).}
  \label{tab:fidelity}
\end{table}

The user message contains one sentence that sets how closely the program should follow the strokes; our application offers three levels, and all results above use \emph{balanced}, the default we fixed before running the benchmark (``follow the strokes' overall size, proportions and where each part sits closely, but clean up and refine the shapes''). \emph{Strict} instead asks the LLM to match positions, proportions and outlines to within about 1\,cm and not to add anything the strokes do not show. We reran 80 sketches with this sentence and nothing else changed (\cref{tab:fidelity}). Strict raises the F-score from 71.1 to 75.8 (higher on 62 of 80 sketches, $p<10^{-7}$), with the largest gains on cabinets (+8.2) and chairs (+7.2). Against the trained model on the same sketches, strict is better overall (75.8 vs.\ 70.7, $p=0.001$) and on airplanes, chairs and tables, has a lower median Chamfer distance (2.41 vs.\ 2.56), is not significantly different after alignment (78.4 vs.\ 77.7, $p=0.45$), and narrows the gap in normal consistency (87.0 vs.\ 89.3). The cabinet deficit shrinks from 12.8 to 4.6 points and is no longer significant ($p=0.11$): strict cabinets have about a fifth less surface area than balanced ones on the same sketches, \ie, fewer interior parts (\cref{sec:cabinets}). The drawn strokes also lie closer to the strict surfaces (median 0.45\% vs.\ 0.59\% of the sketch size). Strict outputs are longer, so a shape costs about 23\% more. Because we chose to test strict after seeing the balanced results, we then ran it, unchanged, on the other 320 test sketches, none of which was used to develop the instruction, the prompt or the check (\cref{s:further}). All 320 programs built, and strict scores 74.1 against 71.3 for balanced and 69.5 for the trained model; it is better than balanced on 243 of 320 sketches ($p<10^{-19}$) and than the trained model on 214 ($p<10^{-8}$), with gains on airplanes, chairs and tables, while cabinets still trail (54.4 vs.\ 61.5); after alignment it remains ahead of the trained model (76.2 vs.\ 74.7, $p=0.001$). We keep balanced for all main results; the held-out run confirms that the fidelity instruction is a strong lever and that those numbers are conservative.

\subsection{Checking against the sketch}
\label{sec:verify_results}

\begin{table}[t]
  \centering
  \footnotesize
  \setlength{\tabcolsep}{3pt}
  \begin{tabular}{@{}lcccccc@{}}
    \toprule
    & Air. & Cab. & Chair & Table & All & + / -- \\
    \midrule
    Balanced & 90.0 & 57.0 & 68.8 & 75.2 & 72.7 & \\
    \quad + check and repair & 91.4 & 58.2 & 70.3 & 75.1 & 73.8 & 18 / 1 \\
    Strict & 92.7 & 66.4 & 75.4 & \textbf{75.9} & 77.6 & \\
    \quad + check and repair & \textbf{93.8} & 66.6 & \textbf{76.1} & 75.8 & \textbf{78.1} & 6 / 4 \\
    \textit{Trained} & \textit{88.1} & \textit{\textbf{67.9}} & \textit{67.1} & \textit{63.1} & \textit{71.6} & \\
    \bottomrule
  \end{tabular}
  \caption{\textbf{Checking against the sketch} (\cref{sec:verify}) on 60 held-out sketches: the 80 of \cref{tab:fidelity} without the 20 used to develop the check. F-score; ``+ / --'': sketches whose F-score the repair raised / lowered. Repair versus no repair: $p=0.001$ (balanced), $p=0.17$ (strict).}
  \label{tab:verify}
\end{table}

We developed the check on the first five sketches per category of the 80 in \cref{tab:fidelity} and evaluated it on the other 60. We added the in-plane condition and the area-based score of \cref{sec:verify} after a development case: with support based on proximity alone and a score based on shares, a repair that deleted a cabinet's drawn doors and back panel raised the score (F-score 61.5 to 50.6). Without ground truth, the check score (with $A_0$ the model's own area) has a Spearman correlation of 0.74 with the F-score on the 60 held-out balanced outputs (0.58 for strict). The particular score is not critical: measures that compare strokes and surface in both directions rank the outputs about as well or better (0.84 vs.\ 0.74 on the 380 balanced outputs not used for development), one-step threshold changes keep our score between 0.66 and 0.80 there, and among the acceptance rules we compared ours lowers the F-score of the fewest balanced sketches, tied with the share-based score (\cref{app:checkdesign}).

\begin{figure}[t]
  \centering
  \includegraphics[width=\linewidth]{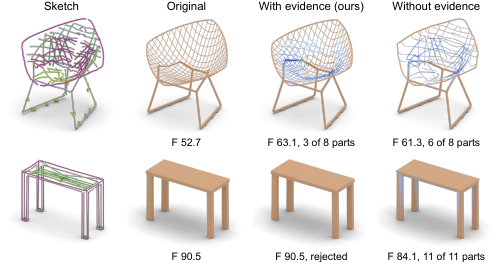}
  \caption{\textbf{Repair with and without the evidence} on two held-out sketches, colored by distance to the original output (unchanged color below 1\% of the object size, light to dark blue from 1\% to 5\%). Chair: with the evidence, the LLM changes 3 of 8 parts, mostly the seat mesh; with only the strokes, it redraws 6 of 8. Table: the check flags the output, and the repair with the evidence changes two aprons but is rejected because the check score does not rise; with only the strokes, the LLM rewrites all 11 parts, the check score rises, and the kept round lowers the F-score from 90.5 to 84.1.}
  \label{fig:repair}
\end{figure}

On the held-out sketches (\cref{tab:verify}), the check flags 32 of 60 balanced models and keeps a repair for 19. Repair raises the F-score from 72.7 to 73.8 (higher on 18 sketches, lower on 1, $p=0.001$), with significant gains on cabinets ($p=0.03$) and chairs ($p=0.01$). The gain is not specific to the official metric: after alignment the F-score also rises (76.0 to 77.1, higher on 17 sketches, lower on 2, $p<10^{-3}$) and the Chamfer distance falls (3.32 to 3.04, $p=0.003$), while normal consistency does not drop (85.3 vs.\ 85.8, $p=0.33$). Of the 25 kept edits, 21 reshape or move parts and 4 delete parts, all interior shelves, drawers or dividers of cabinets that the user did not draw. A kept repair changes a median of 2 of 11 parts; all other parts are rebuilt bit-identically. Repair helps where the check finds problems: balanced outputs with more than the median share of unsupported surface (5.9\%) gain 1.6 points ($p=0.01$), the others 0.5, and a kept repair gains 3.2 points on average. With the strict instruction, the outputs already follow the strokes and have half as much unsupported surface (4.4\% vs.\ 8.6\%): the check flags 20 models, keeps 10 repairs, and the gain (77.6 to 78.1) is not significant ($p=0.17$). The check and the strict instruction thus address the same failure, extra geometry the sketch does not show, once after and once before generation. Checking and repairing costs \$0.33 per flagged sketch on average, \$0.17 per sketch over all 60 balanced sketches (\$0.12 for strict); all parameters, the development results and the other metrics are in \cref{app:checkdesign}. The particular acceptance score matters little: running the same loop end to end but keeping a round when a simpler score improves, the mean distance between strokes and surface in both directions or the F-score against the strokes, gives 74.4 and 74.3, not significantly different from ours ($p=0.16$ and $0.92$; \cref{s:further}). The gain thus does not hinge on our score. Nor does it need the evidence itself: sending only the strokes with the request to fix the model, under the same trigger and acceptance rule, gives a similar mean (73.9, $p=0.53$ vs.\ ours), but lowers the F-score of 9 sketches instead of 1 and changes a median of 6.5 of 11 parts in a kept round instead of 2 (\cref{fig:repair} and \cref{s:further}). The evidence keeps the repair local. On these 60 sketches, strict with repair outperforms the trained model by 6.5 F-score points (78.1 vs.\ 71.6, higher on 44 of 60, $p<10^{-3}$).

\subsection{3D out: the result lands on the sketch}
\label{sec:registration}

\begin{table}[t]
  \centering
  \small
  \resizebox{\linewidth}{!}{\begin{tabular}{@{}llccc@{}}
    \toprule
    & & \multicolumn{2}{c}{Strokes to surface} & Surface to strokes \\
    \cmidrule(lr){3-4}\cmidrule(l){5-5}
    Method & Registration to the sketch & Median dist.\ $\downarrow$ & Within 2\% $\uparrow$ & Within 2\% $\uparrow$ \\
    \midrule
    \ours{} & none (same frame) & \textbf{0.54\%} & \textbf{96.8\%} & \textbf{45.5\%} \\
    Trained & none (canonical frame) & 37.7\% & 0.0\% & 6.5\% \\
    Trained & bounding-box fit & 0.84\% & 89.0\% & 40.2\% \\
    Trained & bounding-box fit + ICP & 0.97\% & 81.2\% & 38.3\% \\
    \midrule
    \textit{\ours{}} & \textit{ICP (for reference)} & \textit{0.74\%} & \textit{89.4\%} & \textit{41.5\%} \\
    \bottomrule
  \end{tabular}}
  \caption{\textbf{Registration to the input sketch} on all 400 sketches. Distance from the points of the sketch's strokes to the generated surface, in the sketch's own frame, as a percentage of the sketch's longest side: median over points (then over sketches), and the fraction of stroke points within 2\%. Bounding-box fit: center and uniformly scale the output to the sketch's bounding box. ICP: symmetric, translation and uniform scale. Last column: the other direction, the share of points sampled on the generated surface that lie within 2\% of a stroke (strokes resampled every 0.5\% of the sketch size).}
  \label{tab:registration}
\end{table}

Because the program is written in the sketch's own frame, its output needs no registration, though the frame alone does not guarantee that parts are placed correctly. \Cref{tab:registration} measures, in the sketch's frame, how far the drawn strokes lie from the generated surface. Without any alignment, the median stroke point lies 0.54\% of the sketch size from our surface, and 96.8\% of stroke points are within 2\%. The trained model generates in a canonical frame, so its raw output is far from the strokes (37.7\%) and has to be registered before it can replace the sketch. Even after it is centered on the sketch and scaled uniformly to the sketch's bounding box, its median distance is 0.84\% (89.0\% of points within 2\%), and our unaligned output is closer on 370 of 400 sketches ($p<10^{-60}$). In the other direction, 45.5\% of our surface lies within 2\% of a stroke, against 40.2\% for the fitted trained model (higher on 290 of 400 sketches, $p<10^{-21}$; 38.3\% after ICP). The interior of a drawn face is far from its outline for both methods, so neither approaches 100\%; cabinets, whose programs build interiors, are the one category where the trained model is higher (30.7\% vs.\ 29.0\%). A symmetric ICP against the sparse strokes improves neither method, because it pulls the surface toward dense clusters of strokes, so we take the bounding-box fit as the stronger registration of the baseline. This measures agreement with what the user drew, not with the ground truth: it rewards following the strokes, including their imprecision, which in-place modeling in VR would require.

\subsection{Why cabinets are hard}
\label{sec:cabinets}

\begin{figure*}[t]
  \centering
  \includegraphics[width=\linewidth,height=0.78\textheight,keepaspectratio]{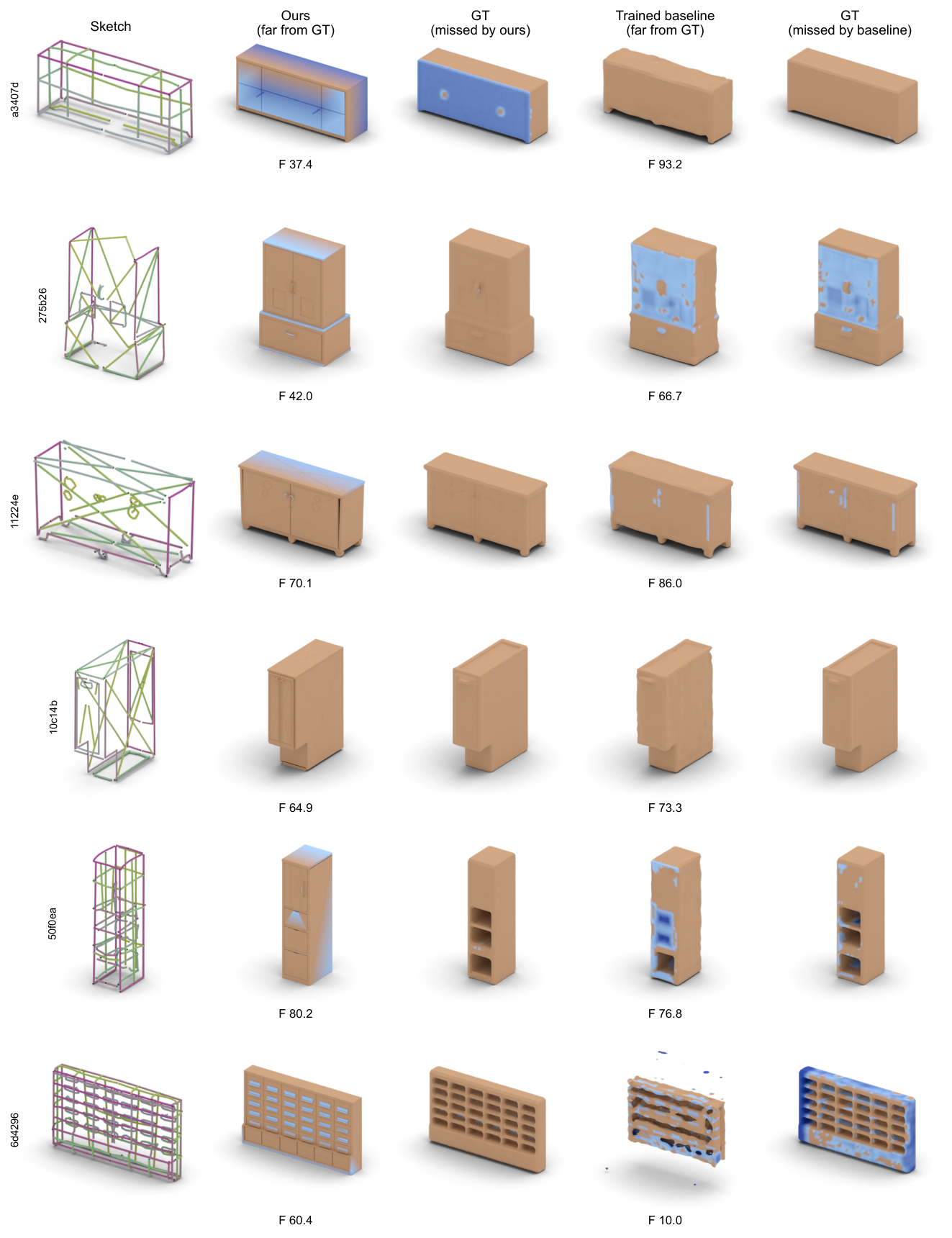}
  \caption{\textbf{Cabinet errors.} Points are colored by their distance to the other surface after the official per-shape normalization (base color: within the F-score threshold $\tau$; light to dark blue: $\tau$ to $3\tau$). Rows are ordered from the largest loss to the largest gain of \ours{} relative to the trained model. The visible surfaces of our cabinets mostly fall within $\tau$; the loss comes from geometry that the solid ground truth lacks (\cref{sec:cabinets}).}
  \label{fig:cabinet}
\end{figure*}

Cabinets are the only category where \ours{} loses, and \cref{fig:cabinet} shows that the visible shape is usually close to the ground truth. The proportions of our cabinets match the ground truth slightly better than those of the trained model (largest normalized side-length error 0.02 vs.\ 0.03), removing small components does not change the score, and alignment does not close the gap (54.5 vs.\ 66.6). Instead, the programs model cabinets as furniture is built, with drawer boxes, shelves, dividers and back panels inside the carcass, while the ground truth is a solid volume with only an outer surface. Our cabinets have a median surface area of 2.50$\times$ that of the ground truth (airplanes 1.21$\times$, chairs 1.29$\times$, tables 1.28$\times$; the trained model 0.99$\times$). Because the protocol samples points uniformly over the whole surface, many of our samples on a cabinet land on interior surfaces far from the ground truth, which lowers precision, and fewer remain on the exterior, which lowers recall. The exposed-surface variant of \cref{tab:more} barely helps here (52.0) because door and drawer seams connect the interior to the outside, so interior surfaces count as exposed. If we first seal gaps of up to about 2.5\% or 5\% of the object size (morphological closing with 3 or 6 voxels at $256^3$) before deciding which points are exposed, only 66\% or 49\% of our cabinet surface remains, and our cabinet F-score rises to 57.7 or 62.3, close to the trained model's 63.0, whose score does not change (63.1 and 63.0). About half of our cabinet surface is therefore interior. A substantial part of the gap thus comes from a mismatch between a part-structured representation and a solid ground truth, although sealing gaps can also hide real geometric errors; the mismatch matters for applications that expect a single closed surface. Asking the LLM to follow the strokes strictly reduces the interior parts and shrinks the cabinet deficit to a non-significant 4.6 points (\cref{sec:fidelity}).

\subsection{Part-level editing}
\label{sec:editing}

\begin{table}[t]
  \centering
  \small
  \setlength{\tabcolsep}{3pt}
  \resizebox{\linewidth}{!}{\begin{tabular}{@{}lccc@{}}
    \toprule
    & Simple (19) & Hard (40) & 3 steps (30) \\
    \midrule
    Untouched parts bit-identical & 161/161 & 338/338 & 249/249 \\
    Edits that changed any untouched part & 0/19 & 0/40 & 0/30 \\
    Target part changed (by $>$\SI{1}{mm}) & 19/19 (19) & 40/40 (34) & 30/30 (30) \\
    Bounding-box size as requested & 17/19 & 38/40 & 26/30 \\
    Target detached from the model & 0/19 & 0/40 & 0/30 \\
    Contacts with other parts kept & 66/66 & 133/137 & 113/120 \\
    Cost / median time per edit & \$0.23 / \SI{29}{s} & \$0.24 / \SI{28}{s} & \$0.26 / \SI{33}{s} \\
    \bottomrule
  \end{tabular}}
  \caption{\textbf{Part-level edits.} Simple: thicker, longer, wider, rounder. Hard: restyle, add details, 40\% bigger, half as thick (\cref{app:edits}). 3 steps: 10 chains of three edits to different parts, each applied to the previous result (thicker, longer, wider, restyle, 40\% bigger, half as thick). Untouched parts are compared as triangle sets after canonical ordering (for 3 steps, once per step, so a part can count up to three times); ``changed'' uses exact point-to-surface distances. Objects are about \SI{1}{m} in size.}
  \label{tab:edit}
\end{table}

We applied 59 single-part edits to generated programs (\cref{tab:edit}): 19 simple geometric edits and 40 harder ones (restyle, add details, 40\% bigger, half as thick) on models not used for the simple set. Because Blender emits the same triangles in a different order when the same script is rebuilt, we compare untouched parts as canonically ordered triangle sets and measure changes with exact point-to-surface distances. Although nothing but the instruction enforces locality, all 499 parts that an edit did not target were rebuilt bit-identically, every target part changed, no edited part detached, and 199 of 203 contacts between the target and its neighbors were kept. Most ``add details'' edits change the part by less than \SI{1}{mm} (grooves and rivets), so we report visible changes separately. Bounding-box ratios match the requested size change in 55 of 59 edits. Edits also compose: in 10 chains of three successive edits to different parts, all 249 untouched parts were rebuilt bit-identically at every step, including the 30 parts edited at an earlier step, and after the third edit all 63 parts never targeted were identical to the original.

\paragraph{Do edits do what was asked?}
\label{sec:edit_judge}
Bounding-box ratios cannot tell whether a restyled part looks more ornate or whether details were added. We therefore had each of the 59 single-part edits judged by two vision-language models from vendors other than the one whose model made the edits, Claude Fable 5.2 and Gemini 4 Argon. Each judge sees one image with the object before and after the edit, in the same framing (two views of the whole object, a close-up of the target in place and the target alone), the target part in blue and all others in gray, together with the instruction, and answers whether the instruction was carried out (yes, partly, no) and whether any gray part changed (\cref{app:edit_judge}). Two controls measure the judges themselves: the same images with an instruction that was not carried out (the opposite size change, or ``40\% bigger'' for edits that keep the size), and the unchanged object shown as both before and after. \Cref{tab:edit_judge} shows the verdicts. The judges find 44 and 45 of the 59 edits done and 7 and 5 partly done, and give the same verdict on 54; in the controls they find at most 1 of 59 done and at most 4 partly done, so only ``done'' is counted as success. Size edits are almost all judged done, restyles done or partly done, and ``add details'' is the hardest kind: the judges find 5 and 7 of the 10 detail edits not done, which matches the geometry (most of these edits change the part by less than \SI{1}{mm}, too little to see). The judges see a gray part change in at most 2 of 59 edits, whereas all untouched parts were bit-identical, a measure of their own false alarms.

\begin{table}[t]
  \centering
  \footnotesize
  \setlength{\tabcolsep}{3pt}
  \resizebox{\linewidth}{!}{\begin{tabular}{@{}lcccc@{}}
    \toprule
    & \multicolumn{2}{c}{Claude Fable 5.2} & \multicolumn{2}{c}{Gemini 4 Argon} \\
    \cmidrule(lr){2-3}\cmidrule(l){4-5}
    & done & partly & done & partly \\
    \midrule
    Edits as performed (59) & \textbf{44} & 7 & \textbf{45} & 5 \\
    \quad thicker, longer, wider, rounder (19) & 16 & 1 & 16 & 1 \\
    \quad restyle (10) & 7 & 3 & 7 & 3 \\
    \quad add details (10) & 2 & 3 & 2 & 1 \\
    \quad bigger, half as thick (20) & 19 & 0 & 20 & 0 \\
    \midrule
    \textit{Control: other instruction} (59) & 0 & 1 & 1 & 1 \\
    \textit{Control: no change} (59) & 0 & 4 & 0 & 1 \\
    \bottomrule
  \end{tabular}}
  \caption{\textbf{Edits judged by two VLMs} from other vendors. ``Done'' and ``partly'': number of edits for which the judge answered that the instruction was carried out or partly carried out; the rest were judged not done. Controls: the same image pairs with an instruction that was not carried out (other instruction), or with the unchanged object as both before and after (no change); the correct answer is ``no''.}
  \label{tab:edit_judge}
\end{table}

\paragraph{Editing through the sketch.}
The trained model has no parts, so the only way to change a part is to change the sketch and regenerate. We simulated this for the 30 of the 32 pure rescalings (bigger, thinner, thicker, longer, wider) whose target part could be matched to strokes: each stroke is assigned to the part whose surface is nearest to most of its points, and the target's strokes are scaled about the part's bounding-box center exactly as the edit requests. We then regenerate with the trained model, either with the original noise seed (the most favorable case) or with a new seed (what a user would get), and measure how much of the surface outside the (generously enlarged) target region moves by more than 1\% of the object size.

\begin{table}[t]
  \centering
  \small
  \setlength{\tabcolsep}{3pt}
  \resizebox{\linewidth}{!}{\begin{tabular}{@{}lccc@{}}
    \toprule
    & \multicolumn{2}{c}{Outside surface moved} & Edits with \\
    & $>$1\% & $>$2\% & leakage \\
    \midrule
    \ours{}: edit the part's code & \textbf{0.0\%} & \textbf{0.0\%} & \textbf{0/30} \\
    Trained: re-sketch, same seed & 23.2\% & 12.7\% & 14/30 \\
    \quad after ICP alignment & 11.5\% & 6.1\% & 13/30 \\
    Trained: re-sketch, new seed & 32.4\% & 16.8\% & 30/30 \\
    \quad after ICP alignment & 23.4\% & 9.6\% & 29/30 \\
    \midrule
    \textit{Trained: no edit, new seed} & \textit{13.2\%} & \textit{3.6\%} & \textit{29/30} \\
    \bottomrule
  \end{tabular}}
  \caption{\textbf{Edit locality vs.\ re-sketching} on 30 rescaling edits. Fraction of the surface outside the target region (its bounding box enlarged by 10\% of the object size) that moves by more than 1\% or 2\% of the object size; an edit ``leaks'' if more than 1\% of that surface moves by more than 1\%. We call an edit that does not leak ``ignored'' if, inside the target region as well, less than 1\% of the surface before or after the edit lies more than 1\% of the object size from the other surface.}
  \label{tab:resketch}
\end{table}

\begin{figure*}[t]
  \centering
  \includegraphics[width=\linewidth,height=0.78\textheight,keepaspectratio]{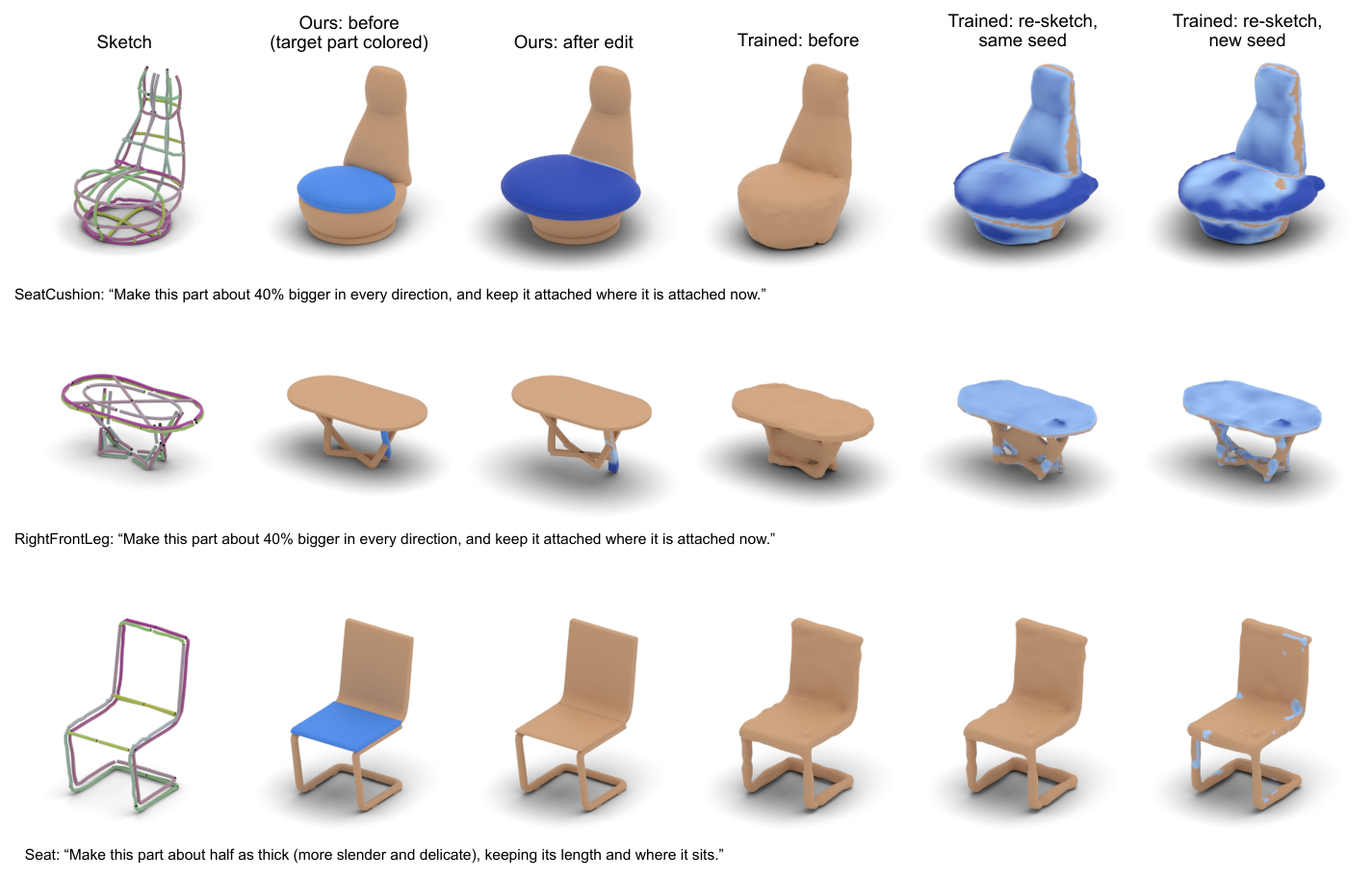}
  \caption{\textbf{Editing a part vs.\ re-sketching.} Columns 2--3: our program before the edit (target part in blue) and after (colored by distance to the original surface; unchanged color below 1\% of the object size, light to dark blue from 1\% to 5\%). Columns 4--6: the trained model on the original sketch, and after scaling the target's strokes and regenerating with the same or a new noise seed, colored the same way. Top: the whole chair changes. Middle: enlarging one leg changes the table top. Bottom: the trained model ignores the thinner seat.}
  \label{fig:edit}
\end{figure*}

\Cref{tab:resketch} shows the result. With the same seed, 14 of 30 re-sketched edits move a substantial part of the rest of the shape (23.2\% of the outside surface on average over all 30 edits; 11.5\% after removing global translation and scale), and the change is not concentrated on the target: the trained model's output changes as much inside the target region (22.4\%) as outside it (23.2\%). In the other 16 edits the rest of the shape stays put, but in 13 of them not even the target region changes (less than 1\% of its surface moves by more than 1\% of the object size): the edit is ignored, typically when the target part is small or thin (9 of the 10 ``half as thick'' edits). Only 3 edits produce a change confined to the target region. With a new seed, every edit changes the rest of the shape, and regenerating the \emph{unedited} sketch with a new seed already moves 13.2\% of the surface. Editing the program leaves the rest of the shape untouched in all cases. \Cref{fig:edit} shows examples. This locality comes from having parts rather than from the program as such: in a mesh perfectly segmented into our own parts, directly scaling the target, about its center or about its contacts, also leaves the rest untouched, meets the requested size in all 30 edits (ours 27) and keeps 98 or 100 of 107 contacts with neighboring parts (ours 103; \cref{s:further}). A transform cannot express the other 27 edits, which make a part rounder, restyle it or add details; the program applies them with all 92 contacts kept, although added details are often too faint to see (\cref{sec:edit_judge}). And the program supplies the segmentation that such a baseline presupposes.

\section{Limitations}
\label{sec:limitations}

\paragraph{Metrics beyond the official F-score.}
The two methods do not differ significantly in the official Chamfer distance or in F-score after ICP alignment, but the trained model has a lower Chamfer distance after alignment and higher normal consistency against the smooth references (\cref{sec:more_metrics}). Our outputs are crisp and add details the ground truth does not have (overhangs, panels, feet, handles, interiors), which these metrics penalize. Whether people prefer them would need a perceptual study.

\paragraph{Proprietary models.}
All main results use one proprietary LLM accessed through an API whose behavior can change over time. The prompts are given verbatim (\cref{app:prompts}), but we cannot guarantee that the same numbers can be reproduced later. Two other proprietary models also reach the trained model's level on the same 60 sketches, one of them slightly below GPT-6 Astra, but the strongest open-weight model we tried reaches 40.6 against 71.0 on 60 sketches (\cref{sec:openllm}), so the approach cannot yet be reproduced with open models alone.

\paragraph{Variance, cost and latency.}
Results vary between runs for individual sketches (\cref{sec:ablation}). Each shape costs about \$0.31 at list prices and takes about a minute, whereas the trained model runs in about \SI{29}{s} on a laptop GPU at no marginal cost.

\paragraph{Prompt.}
The prompt was written for an interactive modeling application and includes rules that do not apply to the benchmark categories. We did not tune it on the benchmark; the official training sketches have not been released, so there is no clean data to tune on. This keeps the comparison honest but probably leaves accuracy on the table, for example on cabinet interiors.

\paragraph{Baselines.}
LL3M~\cite{lu2025ll3m} could not be run, so our agent baseline re-implements its loop without retrieval from the Blender documentation or a verification agent and with one critique round; Nova3D~\cite{noor2026nova3d} was not compared. The 2D detour stands for its route rather than for the strongest system that takes it.

\paragraph{Editing evaluation.}
The edit instructions were generated by us from templates, not collected from users. Whether an edit did what was asked is judged by bounding-box ratios and by two VLMs (\cref{sec:edit_judge}), not by people; the VLMs reject mismatched and unchanged pairs almost always but may still miss subtle failures. The comparison with the trained model simulates re-sketching by scaling the strokes of one part, which covers only edits that are pure rescalings. The multi-step chains edit different parts; repeated edits of one part and edits that require several parts to change together are untested, and we did not check part names against a semantic segmentation.

\paragraph{Scope.}
The benchmark covers four rigid furniture and vehicle categories with one sketch per shape, drawn in a lab. Beyond them, we tested 56 synthetic sketches, 36 of six ModelNet10 categories and 20 of people and plants from ModelNet40, with references we rebuilt ourselves (\cref{sec:ood}); we have not tested scenes or sketches drawn by people under other conditions. Missing components and rotated sketches were tested on 40 sketches only (\cref{sec:serialization}).

\section{Conclusion}
\label{sec:conclusion}
On the four rigid categories of the VRSketch2Shape benchmark, staying in 3D from input to output is enough for a frontier general-purpose LLM to turn VR sketches into part-structured 3D models without any training: writing the strokes down as coordinates and asking for a Blender program that builds named parts in the sketch's frame matches, and on the official F-score exceeds, a state-of-the-art model trained for this task on its own benchmark. On synthetic sketches of four further categories, it stayed ahead of that model, which degraded outside its training categories; it also built people and plants, irregular shapes far from the benchmark's, and matched their raw meshes much better than the trained model and a 2D detour. The 3D coordinates carry almost all of the signal; replacing the coordinates by rendered views loses much of it, a detour through an image generator and an image-to-3D model loses more, and the category name is not needed. The result is registered to the sketch by construction. Unlike an SDF, the output is a program of named parts, and part-level edits expressed as program rewrites left every untouched part bit-identical in our experiments, whereas re-sketching and regenerating with the trained model either ignores the edit or changes the rest of the shape. Placed in the frame where the program builds, the output also follows a rotated sketch, and comparing it with the strokes, without ground truth, points to the parts to repair. Two pipelines of different design and three frontier LLMs reach the trained model's level from the same coordinates, while rendered views of the same strokes do not. We hope these results serve as a strong training-free baseline for sketch-based 3D modeling, and as evidence that the representation, strokes as coordinates in the frame where the program builds, more than a particular prompt or pipeline, is what other methods can reuse.

{
    \small
    \bibliographystyle{ieeenat_fullname}
    \bibliography{refs}
}

\clearpage
\appendix
\section{Checking against the sketch: parameters and further results}
\label{app:checkdesign}

\paragraph{Parameters.} All lengths are in the sketch frame, in which the sketch's longest side measures \SI{100}{cm}. Strokes are resampled every \SI{0.5}{cm}; each model surface is sampled with 120k area-weighted points. A stroke point is covered if it lies within \SI{2}{cm} of the surface; a stroke is missed if more than half of its points are not covered. A surface point is supported if it lies within \SI{3}{cm} of a stroke point, or if, within \SI{60}{cm}, stroke points that lie less than \SI{2}{cm} from its tangent plane, on strokes whose direction makes an angle of more than \ang{60} with the surface normal, are found in all four quadrants around it (computed on 12k surface points and propagated to the others by nearest neighbor). A model is sent for repair if any stroke is missed, or if a part covering at least 3\% of the surface has less than 30\% support or less stroke length assigned to it (stroke points for which it is the nearest part) than its largest extent. The check message lists at most 10 missed strokes, ordered by length times median distance, and at most 15 parts, least supported first (\cref{app:prompts}). A repair is kept if the score of \cref{sec:verify} rises by more than 0.01 and the share of covered stroke points drops by at most 0.01; at most two rounds are run. The edit uses the fidelity sentence of the run being repaired. These values were set before the held-out evaluation, on the 20 development sketches.

\paragraph{Development set.} On the 20 development sketches, repair raised the balanced F-score from 66.1 to 69.3 (10 better, 2 worse) and left the strict F-score unchanged (70.5; 3 better, none worse). The development case that led to the in-plane condition and the area-based score is described in \cref{sec:verify_results}.

\paragraph{Design choices.} \Cref{tab:check_design} compares our score with simpler measures of agreement between strokes and surface and changes each threshold by one step. The particular score is not critical. Measures that look in both directions (the mean distance, or an F-score that treats the strokes as ground truth) rank the outputs about as well as ours on the held-out balanced sketches and better on the strict outputs and on the 380 balanced test outputs; coverage alone is uninformative because almost every output covers the strokes. The in-plane condition improves the ranking on the held-out balanced sketches but worsens it on the strict outputs and on the 380 outputs; we added it for the repair feedback, where proximity alone reports drawn faces such as doors as unsupported (\cref{sec:verify_results}). One-step threshold changes move the correlation by less than 0.1 on the 380 outputs. \Cref{tab:accept} applies other acceptance rules to the same repair candidates: ours and the share-based score lower the F-score of the fewest balanced sketches (1 of 60; on strict outputs 2, against 1 for coverage only), whereas accepting every repair gains slightly more on average than ours but lowers it for 6 balanced and 5 strict sketches.

\begin{table}[t]
  \centering
  \footnotesize
  \setlength{\tabcolsep}{1.8pt}
  \begin{tabular}{@{}lccc@{}}
    \toprule
    Score & Bal.\ 60 & Strict 60 & Bal.\ 380 \\
    \midrule
    Ours, $c + (A_s - A_u)/A_0$ & 0.74 & 0.58 & 0.74 \\
    \quad without the in-plane condition & 0.69 & 0.64 & 0.83 \\
    \quad coverage $c$ only & 0.16 & 0.01 & 0.23 \\
    \quad support $1-u$ only & 0.72 & 0.57 & 0.73 \\
    Mean dist., strokes to surface & 0.42 & 0.40 & 0.53 \\
    Mean dist., both directions & 0.73 & 0.70 & 0.84 \\
    F-score vs.\ strokes (\SI{2}{cm}) & 0.72 & 0.68 & 0.84 \\
    \midrule
    Ours, one threshold $\pm$ one step & 0.65--0.79 & 0.43--0.65 & 0.66--0.80 \\
    \bottomrule
  \end{tabular}
  \caption{\textbf{Ranking outputs without ground truth}: Spearman correlation of each score with the F-score. Bal.\ 60, Strict 60: the held-out sketches; Bal.\ 380: all balanced test outputs except the 20 development sketches. For unrepaired outputs $A_0$ is the output's own area, so our score is $c + 1 - 2u$ with $u$ the unsupported share. Threshold steps: coverage 1, 3 or \SI{4}{cm} (default 2), support 2 or \SI{5}{cm} (3), in-plane radius 30 or \SI{90}{cm} (60), plane band 1 or \SI{3}{cm} (2).}
  \label{tab:check_design}
\end{table}

\begin{table}[t]
  \centering
  \footnotesize
  \setlength{\tabcolsep}{3pt}
  \begin{tabular}{@{}lcccc@{}}
    \toprule
    & \multicolumn{2}{c}{Balanced} & \multicolumn{2}{c}{Strict} \\
    \cmidrule(lr){2-3}\cmidrule(l){4-5}
    Acceptance rule & F $\uparrow$ & + / -- & F $\uparrow$ & + / -- \\
    \midrule
    No repair & 72.9 & & 77.6 & \\
    Ours (area score, coverage $-$0.01 at most) & 73.9 & 13 / 1 & 78.1 & 6 / 2 \\
    Share-based score & 74.3 & 11 / 1 & 78.1 & 6 / 2 \\
    Coverage only & 73.6 & 7 / 3 & 78.0 & 4 / 1 \\
    Accept every repair & 74.2 & 17 / 6 & 78.1 & 9 / 5 \\
    \bottomrule
  \end{tabular}
  \caption{\textbf{Acceptance rules} applied to the stored repair candidates of the 60 held-out sketches. F-score as computed in the repair step (within 0.2 of the main tables); ``+ / --'': sketches whose F-score changed by more than 0.5. The candidates were generated with our feedback, and a second round exists only where our rule accepted the first.}
  \label{tab:accept}
\end{table}

\paragraph{What the repairs change.} On the 60 held-out sketches, the check flagged 32 balanced models and 20 strict models; 19 and 10 of them kept at least one repair (25 and 13 kept rounds). Kept repairs changed a median of 2 parts (of 11 and 10); all other parts were rebuilt bit-identically. Four kept balanced edits deleted parts (9 in total, all interior shelves, drawers or dividers of cabinets) and strict edits deleted 2; none added a part. The score rewards added supported surface, which a repair could exploit by adding geometry along the strokes. Over the 80 sketches, 45 outputs kept at least one repair (31 balanced and 14 strict; 29 held-out and 16 development outputs, all from the final pipeline; 58 kept rounds in total). Comparing each final output with the original, the total surface area changed by a median of +0.9\% (at most +11\%). Each of the 12 outputs whose area grew by more than 2\% also covered more stroke points (by 2 to 14 percentage points), and their F-score changed by a median of +1.7 (9 higher, 3 lower, by at most 3.2). The area term thus did not drive the repairs. The mean cost per sketch, including sketches that were not flagged, was \$0.17 (balanced) and \$0.12 (strict).

\begin{table}[t]
  \centering
  \footnotesize
  \setlength{\tabcolsep}{3pt}
  \begin{tabular}{@{}lcccc@{}}
    \toprule
    & F $\uparrow$ & F al.\ $\uparrow$ & CD al.\ $\downarrow$ & NC $\uparrow$ \\
    \midrule
    Balanced & 72.7 & 76.0 & 3.32 & 85.3 \\
    \quad + check and repair & \textbf{73.8}$^{**}$ & \textbf{77.1}$^{***}$ & \textbf{3.04}$^{**}$ & 85.8 \\
    Strict & 77.6 & 80.5 & 2.41 & 87.2 \\
    \quad + check and repair & 78.1 & 81.0 & 2.36 & 87.3 \\
    \bottomrule
  \end{tabular}
  \caption{\textbf{Repair on the 60 held-out sketches}, official and aligned metrics. Stars: paired Wilcoxon test against no repair, $^{**}p<0.01$, $^{***}p<0.001$; NC and all strict differences are not significant ($p\geq0.11$).}
  \label{tab:repair_metrics}
\end{table}

\paragraph{Not only the official metric.} \Cref{tab:repair_metrics} gives the official and aligned metrics for both instructions. On balanced outputs the gain from repair also appears after alignment (\cref{sec:verify_results}), while normal consistency does not change significantly; on strict outputs no metric changes significantly.

\section{Unseen categories, people and plants}
\label{s:ood}
The benchmark has four categories, and the trained model was trained on all four. To test other categories we made sketches from ModelNet10 test shapes~\cite{wu2015shapenets}. Bed, monitor, toilet and bathtub differ from all four training categories (we left out desk, dresser, night stand and sofa, which resemble table, cabinet and chair); chair and table are a control the trained model has seen, so that the effect of the synthetic drawing style can be told apart from that of the category. We drew six shapes per category from the test split at random (fixed seed) and rotated each category to the benchmark's convention: front toward the viewer, checked on chair backs, toilet tanks, bed heads and monitor screens; tables and bathtubs, which have no front, with the long side in front, as for most benchmark tables.

\begin{figure}[t]
  \centering
  \includegraphics[width=\linewidth]{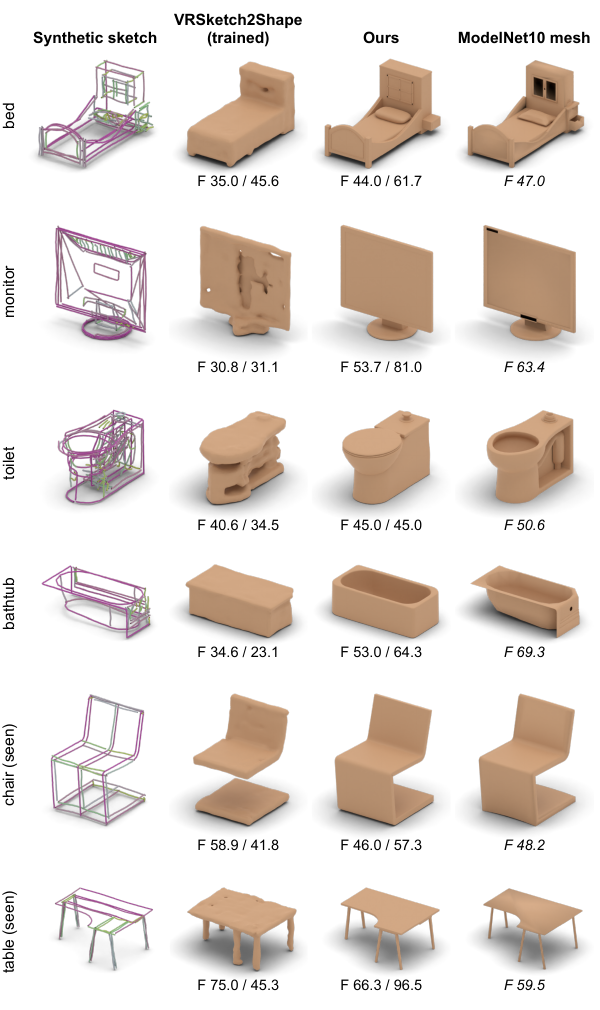}
  \caption{\textbf{Unseen categories}: one sketch per category, the one at the median F-score difference between \ours{} and the trained model (not selected for appearance). Below each output: F@0.04 against the rebuilt reference / against the raw mesh; below the mesh, in italics: the mesh itself against the rebuilt reference. Chair and table are categories the trained model has seen. All shapes are normalized to their own bounding box; dark areas on the meshes are flipped faces of the ModelNet10 files.}
  \label{fig:ood}
\end{figure}

\paragraph{Sketch synthesis.} The sketch generator of~\cite{chen2026vrsketch2shape} is not public, so we wrote our own. After removing duplicated faces (many ModelNet10 meshes are double-sided), it takes crease edges (dihedral angle above \ang{35}), open boundaries and the silhouettes seen along the three axes, and chains them into polylines, continuing through junctions where the direction changes by less than \ang{30} and splitting where a line turns back by more than \ang{120}. Lines are simplified (tolerance 0.5\% of the object size); lines shorter than 3\%, lines that fold back on themselves and lines that mostly retrace a longer one (70\% of their points within 1.5\% of it) are dropped, and at most 90 strokes are kept, longest first. Each stroke is resampled every 2.5\%, extended or shortened by up to 2\% at each end, and given a smooth wobble (0.2--0.4\%) and point noise (0.05\%). Meshes that yield fewer than 15 strokes were skipped (one bathtub). The sketches have a median of 70 strokes (80 sampled benchmark sketches: 79) but are sparser: on average 67\% of the reference surface lies within 5\% of a stroke, against 83\% for the benchmark sketches. \Cref{fig:ood} shows examples.

\paragraph{Methods.} Both methods receive the same sketches. The trained model runs exactly as on the benchmark (same stroke simplification, normalization and seed per sketch). \ours{} uses the default input (coordinates, views and the category name, \eg ``a bed'') and the same prompt. All 36 programs built (median 8 parts; \$10.61 in total).

\paragraph{Reference surfaces.} The benchmark's references are the 0.02 level set of a $64^3$ SDF, a surface offset slightly outward from the shape, and the trained model was trained on such SDFs. We rebuilt references for ModelNet10 the same way: a distance field on the same grid, signed by flood fill from outside, and its 0.02 level set (median offset 1.50\% of the object size; benchmark: 1.45\%). F@0.04 is sensitive to offsets of this size, and the two methods sit on different sides of it: our parts lie on the shape's surface, while the trained model reproduces the offset SDF. A perfect reconstruction shows the effect: the ModelNet10 mesh itself scores only 53.1 on the control categories and 52.2 on the unseen ones against the rebuilt references (\cref{tab:ood}), and the trained model scores above it on 9 of the 12 control sketches. Against the raw meshes, the ranking on the control categories reverses (\ours{} 77.7, trained 46.0). The two control examples in \cref{fig:ood} show both: our chair and table are close to the meshes, and the trained model scores higher than the meshes themselves against the rebuilt references. Parts our programs miss do not explain the control gap: the share of the reference surface farther than 5\% of the object size from the output, \ie missing parts rather than offsets, is 3.3\% for \ours{} and 5.7\% for the trained model (raw meshes: 2.1\% and 7.3\%). We therefore report both references and compare the methods where the ranking does not depend on the reference.

\paragraph{Results.} On the unseen categories, \ours{} scores higher than the trained model on 20 of 24 sketches against the rebuilt references (50.4 vs.\ 38.2, $p<0.001$, paired Wilcoxon) and on 23 of 24 against the raw meshes (65.5 vs.\ 36.8, $p<0.001$); the difference is also significant at F@0.06, F@0.1 and in Chamfer distance under both references. From the control to the unseen categories (\cref{tab:ooddrops}), the trained model falls from 66.2 to 38.2 against the rebuilt references ($p<0.001$), well below the 52.2 of the meshes themselves, while \ours{} goes from 53.1 to 50.4 ($p=0.58$); the drops differ ($p<0.001$), also at the looser thresholds and in Chamfer distance. Against the raw meshes, both lose about ten points at F@0.04 (trained 46.0 to 36.8, \ours{} 77.7 to 65.5), and the drops differ only at F@0.1 (trained 92.0 to 80.0, \ours{} 98.2 to 95.2, $p=0.02$). On the unseen categories the trained model returns shapes that fill the extent of the strokes without their structure, while \ours{} returns recognizable objects (\cref{fig:ood}).

\paragraph{A baseline not trained on any category.} The trained model's drop outside its categories could by itself make \ours{} look good. We therefore also ran the 2D detour of \detourref, which uses general-purpose image and image-to-3D models trained on no particular category, on the same 36 sketches, unchanged: the same three-quarter rendering of the strokes, the same prompt template with the category name, and the best of 24 rotations per sketch. It scores 27.4 on the unseen and 36.0 on the control categories against the rebuilt references, and 31.2 and 46.5 against the raw meshes (\cref{tab:ood}). \ours{} is better than the detour on 23 of the 24 unseen sketches against the rebuilt references ($p<10^{-5}$) and on all 24 against the raw meshes ($p<10^{-6}$), and also on the control categories (9 and 12 of 12). The detour's Chamfer distance on the unseen categories is four times ours (28.6 vs.\ 7.0, rebuilt references). Like the trained model, the detour drops from the control to the unseen categories (36.0 to 27.4 and 46.5 to 31.2; not significant, $p=0.20$ and $p=0.11$), so part of every method's drop comes from these four categories and their synthetic sketches, not only from training. Its failures resemble those on the benchmark: the image model often turns hidden and silhouette lines into wire frames, rails and glass panes around the object (\eg cages around toilets and beds), which the image-to-3D model then builds, although some of its toilets and bathtubs are clean.

\begin{figure*}[t]
  \centering
  \includegraphics[width=\textwidth]{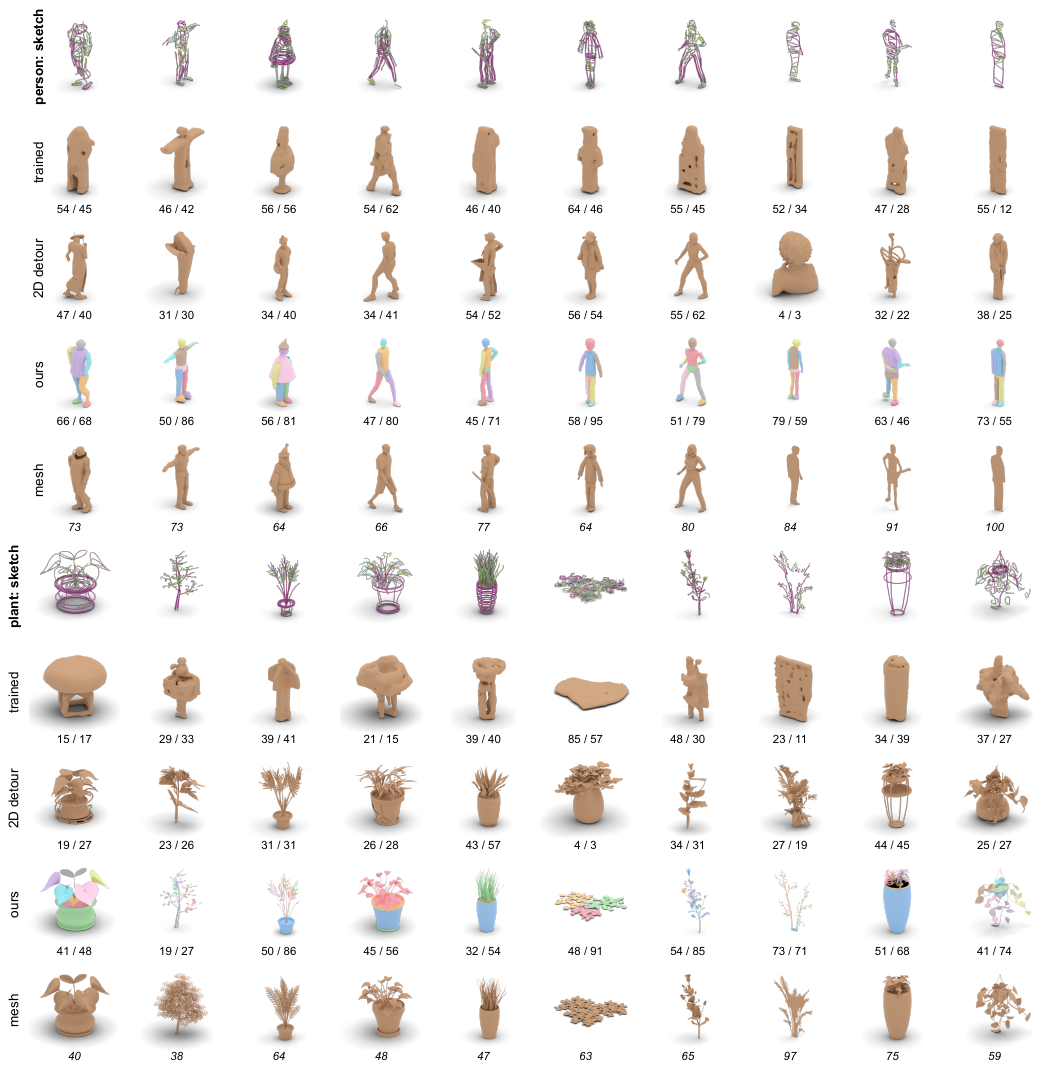}
  \caption{\textbf{All 20 people and plants}, in the order they were drawn from ModelNet40: synthetic sketch, the trained model, the 2D detour, \ours{} (one color per part) and the ModelNet40 mesh. Below each output: F@0.04 against the rebuilt reference / against the raw mesh; below the mesh, in italics: the mesh itself against the rebuilt reference.}
  \label{fig:organic_all}
\end{figure*}

\paragraph{People and plants.} The categories above are rigid objects. To test irregular shapes, we took ten test shapes each of person and plant from ModelNet40~\cite{wu2015shapenets}, again at random with a fixed seed (one person gave fewer than 15 strokes and was replaced by the next), and synthesized sketches with the same generator. Their many small features (faces, hands, leaves) give every sketch the maximum of 90 strokes, and the sketches cover the reference more densely than those above: on average 92.5\% of the surface lies within 5\% of a stroke (people 98.9\%, plants 86.1\%). All three methods ran unchanged with the category name (``a person'', ``a plant''), and the references were rebuilt as above. Our prompt, written earlier for our modeling application and not changed for this test, contains general rules for people (limbs as continuous tubes, clothing, hands and faces; \promptsref) but none for plants. All 20 programs built (median 10 parts; \$9.39 in total). They build people from a head, torso, arms and legs that follow the sketched pose, and plants from pots, stems and leaves modeled one by one (\cref{fig:organic_all}).

\begin{figure}[t]
  \centering
  \includegraphics[width=\linewidth]{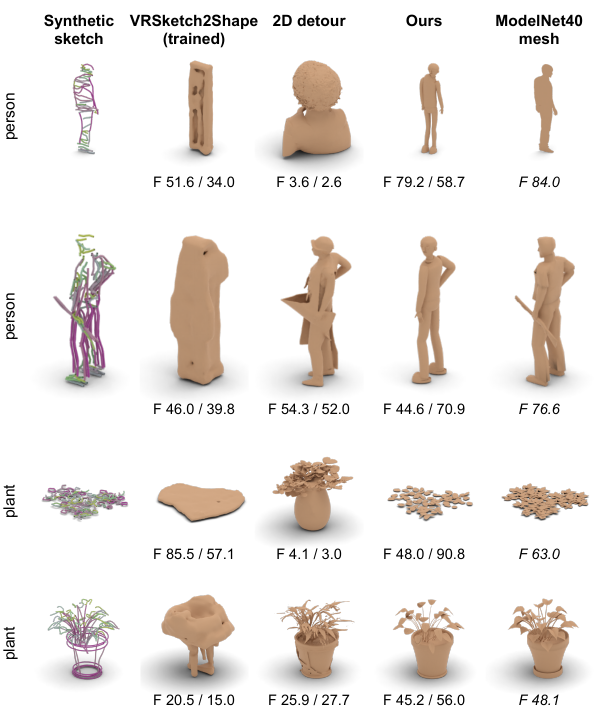}
  \caption{\textbf{People and plants}: per category, the two sketches at the median F-score difference between \ours{} and the trained model against the raw meshes (not selected for appearance). Below each output: F@0.04 against the rebuilt reference / against the raw mesh; below the mesh, in italics: the mesh itself against the rebuilt reference. The 2D detour is shown at the rotation chosen against the raw mesh. The ModelNet40 mesh of the top person is a flat cut-out. All shapes are normalized to their own bounding box.}
  \label{fig:organic}
\end{figure}

Against the raw meshes, \ours{} scores 68.9, the trained model 36.0 and the 2D detour 33.1, and \ours{} is better than each on 19 of the 20 sketches ($p<10^{-5}$), for people and plants alike (\cref{tab:organic}); plants, for which the prompt has no specific rules, show the same advantage (65.9 vs.\ 30.9). Against the rebuilt references, \ours{} is still better than the detour (52.0 vs.\ 33.0, 16 of 20, $p<0.001$), but its lead over the trained model (52.0 vs.\ 45.0, 12 of 20) is not significant ($p=0.10$). These references favor thick, smooth outputs even more than on ModelNet10: the offset turns thin limbs, stems and leaves into slabs, and the meshes themselves score only 68.4 against them (plants 59.8). The ground-cover plant in \cref{fig:organic} shows the effect: the trained model's flat slab scores 85.5 against the rebuilt reference, above the mesh itself (63.0), but 57.1 against the raw mesh, on which our separate leaves score 90.8. Elsewhere the trained model returns blobs that fill the extent of the strokes. The detour's image model draws a plausible person or plant, but often not the sketched one: a portrait instead of a full figure, props that are not in the sketch such as sheets of cardboard, pot stands as wire frames, in two cases an abstract sculpture, and the ground cover in a vase.

\paragraph{Limits.} These are 56 sketches from one synthetic generator, six or ten per category, and references we rebuilt ourselves. They show that the method is not limited to the benchmark's categories or to rigid shapes and that the trained model degrades outside them; they do not show how either does with sketches drawn by people.

\begin{table}[t]
  \centering
  \footnotesize
  \setlength{\tabcolsep}{2.2pt}
  \resizebox{\linewidth}{!}{\begin{tabular}{@{}lcccccccc@{}}
    \toprule
    & \multicolumn{5}{c}{Unseen by the trained model} & \multicolumn{3}{c}{Seen (control)} \\
    \cmidrule(lr){2-6}\cmidrule(l){7-9}
    Method, reference & Bed & Mon. & Toil. & Bath. & All (24) & Chair & Table & All (12) \\
    \midrule
    Trained, benchmark-style & 39.4 & 32.7 & 41.2 & 39.5 & 38.2 & \textbf{54.8} & \textbf{77.5} & \textbf{66.2} \\
    \ours{}, benchmark-style & \textbf{46.3} & \textbf{55.2} & \textbf{41.6} & \textbf{58.4} & \textbf{50.4} & 40.6 & 65.6 & 53.1 \\
    2D detour, benchmark-style & 34.1 & 26.8 & 23.8 & 25.0 & 27.4 & 26.2 & 45.8 & 36.0 \\
    \textit{Mesh itself}, benchmark-style & \textit{51.9} & \textit{50.2} & \textit{50.8} & \textit{55.9} & \textit{52.2} & \textit{50.6} & \textit{55.6} & \textit{53.1} \\
    \midrule
    Trained, raw mesh & 42.1 & 32.4 & 38.0 & 34.7 & 36.8 & 42.5 & 49.5 & 46.0 \\
    \ours{}, raw mesh & \textbf{67.3} & \textbf{79.7} & \textbf{51.9} & \textbf{63.2} & \textbf{65.5} & \textbf{62.7} & \textbf{92.7} & \textbf{77.7} \\
    2D detour, raw mesh & 46.1 & 26.2 & 28.5 & 24.0 & 31.2 & 34.9 & 58.1 & 46.5 \\
    \bottomrule
  \end{tabular}}
  \caption{\textbf{Unseen categories} (F@0.04, 6 synthetic sketches per category). ``Benchmark-style'': reference surface rebuilt like the benchmark's (offset surface of a $64^3$ distance field); ``raw mesh'': the ModelNet10 surface itself. 2D detour: at its best rotation per sketch. ``Mesh itself'': the ModelNet10 mesh scored against the benchmark-style reference, \ie what a perfect reconstruction scores; on the seen categories the trained model scores above it by reproducing the reference's offset (see text). Bold: the best method per column and reference.}
  \label{tab:ood}
\end{table}

\begin{table}[t]
  \centering
  \footnotesize
  \setlength{\tabcolsep}{2.5pt}
  \resizebox{\linewidth}{!}{\begin{tabular}{@{}llccccc@{}}
    \toprule
    & & \multicolumn{2}{c}{Trained} & \multicolumn{2}{c}{\ours{}} & Drops \\
    \cmidrule(lr){3-4}\cmidrule(lr){5-6}
    Reference & Metric & seen\,$\to$\,unseen & $p$ & seen\,$\to$\,unseen & $p$ & differ, $p$ \\
    \midrule
    Benchmark-style & F@0.04 & 66.2\,$\to$\,38.2 & $<$0.001 & 53.1\,$\to$\,50.4 & 0.580 & $<$0.001 \\
     & F@0.06 & 84.8\,$\to$\,63.3 & $<$0.001 & 80.1\,$\to$\,79.4 & 0.750 & $<$0.001 \\
     & F@0.1 & 94.2\,$\to$\,80.0 & $<$0.001 & 94.8\,$\to$\,93.3 & 0.274 & 0.005 \\
     & CD$\times10^3$ & 7.7\,$\to$\,18.7 & 0.001 & 5.7\,$\to$\,7.0 & 0.247 & 0.009 \\
    \midrule
    Raw mesh & F@0.04 & 46.0\,$\to$\,36.8 & 0.016 & 77.7\,$\to$\,65.5 & 0.046 & 0.450 \\
     & F@0.06 & 71.5\,$\to$\,60.0 & 0.010 & 91.7\,$\to$\,86.6 & 0.084 & 0.392 \\
     & F@0.1 & 92.0\,$\to$\,80.0 & 0.002 & 98.2\,$\to$\,95.2 & 0.041 & 0.022 \\
     & CD$\times10^3$ & 10.1\,$\to$\,21.0 & 0.015 & 3.1\,$\to$\,5.5 & 0.054 & 0.097 \\
    \bottomrule
  \end{tabular}}
  \caption{\textbf{From seen to unseen categories}, mean over the 12 control and 24 unseen sketches. $p$: Mann--Whitney test of seen vs.\ unseen for each method; last column: whether the per-sketch difference between the methods changes from seen to unseen categories (Mann--Whitney). Lower CD is better.}
  \label{tab:ooddrops}
\end{table}

\begin{table}[t]
  \centering
  \footnotesize
  \begin{tabular}{@{}lccc@{}}
    \toprule
    Method, reference & Person & Plant & All (20) \\
    \midrule
    Trained, benchmark-style & 52.9 & 37.2 & 45.0 \\
    \ours{}, benchmark-style & \textbf{58.6} & \textbf{45.4} & \textbf{52.0} \\
    2D detour, benchmark-style & 38.4 & 27.7 & 33.0 \\
    \textit{Mesh itself}, benchmark-style & \textit{77.0} & \textit{59.8} & \textit{68.4} \\
    \midrule
    Trained, raw mesh & 41.0 & 30.9 & 36.0 \\
    \ours{}, raw mesh & \textbf{71.9} & \textbf{65.9} & \textbf{68.9} \\
    2D detour, raw mesh & 36.9 & 29.4 & 33.1 \\
    \bottomrule
  \end{tabular}
  \caption{\textbf{People and plants} (F@0.04, 10 synthetic sketches each of person and plant from ModelNet40). References, rows and bold as in \cref{tab:ood}. \ours{} is better than the trained model on 12 of 20 sketches against the benchmark-style references ($p=0.10$, paired Wilcoxon) and on 19 of 20 against the raw meshes ($p<10^{-5}$), and better than the 2D detour on 16 ($p<0.001$) and 19 of 20 ($p<10^{-5}$).}
  \label{tab:organic}
\end{table}

\section{Further baselines and robustness tests}
\label{s:further}
This section gives the details of five tests summarized in the main paper: a structured multi-agent baseline with the same LLM and input, the strict instruction on held-out sketches, repair with simpler acceptance scores, editing a segmented mesh, and rotated sketches and sketches missing a whole component. All use the official protocol and paired Wilcoxon tests. The last paragraph, not in the main paper, compares the proportions of the outputs with those of their references.

\paragraph{A structured agent baseline.}
LL3M~\cite{lu2025ll3m} writes Blender code with several agents, but it cannot be run: its public repository contains only a client, the agents ran on the authors' server, which has been shut down, and its coding model (Claude 3.7 Sonnet; GPT-4o and Gemini 2.0 Flash ran the other agents) has been retired. We therefore re-implemented its loop as described in the paper, with GPT-6 Astra for every agent (Responses API, default parameters, as for \ours{}). A \emph{planning} agent decomposes the object into components and gives each one's shape, approximate dimensions, position and connections, using the coordinates, without writing code. A \emph{coding} agent writes a Blender~3.6 script that builds each component as a separate, descriptively named mesh object, without cameras, lights or a ground plane. If the script fails, the coding agent receives the error and fixes it (at most twice). A \emph{critic} then sees four renderings of the result (front, right, top, three-quarter) next to renderings of the sketch strokes from the same views and at the same scale, together with the request, the plan and the script, and lists discrepancies in shape, proportions, position and missing or extra components, or replies that there are none. If it lists any, the coding agent revises the script once (one more fix allowed). Apart from the critic's renderings of the strokes, all agents receive exactly the input of the minimal-prompt test and of our coordinates-only condition: the category name, the sketch size, the same serialized coordinates and the same description of the axes. Our version differs from LL3M in five respects: there is no retrieval from the Blender documentation and no verification agent, the critique runs once, the critic compares the result with the sketch rather than with a text prompt, and one LLM, ours, serves all agents.

On the 40 sketches of the minimal-prompt test (\cref{tab:ll3m}), all 40 scripts built (6 after a fix). The critic found nothing to change for 23 sketches; for the other 17 the revision built and raised the mean F-score from 71.5 to 72.2 ($p=0.03$). The baseline reaches 72.2, not significantly different from \ours{} with coordinates only (70.6, $p=0.18$, each better on 20 of 40; $p=0.47$ before the critique) and from the trained model (69.0, $p=0.24$), and clearly above the minimal prompt (65.9, $p<10^{-3}$). It makes 3.6 LLM calls per sketch and costs \$0.83, against one call (plus a fix in 6 of 400 cases) and \$0.29 for \ours{} with coordinates only, and it returns a median of 41 objects instead of 10 parts. A second pipeline of different design thus reaches a comparable level from the same coordinates: the result rests on giving the LLM the strokes as coordinates in the frame in which it builds, not on our prompt or on a particular agent design. The pipelines differ in cost and in how finely they split the object.

\begin{table}[t]
  \centering
  \footnotesize
  \setlength{\tabcolsep}{4pt}
  \begin{tabular}{@{}lcccc@{}}
    \toprule
    & LL3M-style & \ours{} & Minimal & Trained \\
    \midrule
    Airplane & 92.4 & 89.5 & 91.0 & 89.5 \\
    Cabinet & 62.9 & 61.5 & 58.9 & 61.7 \\
    Chair & 67.7 & 65.5 & 51.5 & 66.8 \\
    Table & 65.7 & 65.9 & 62.0 & 57.9 \\
    \midrule
    All (40) & 72.2 & 70.6 & 65.9 & 69.0 \\
    \bottomrule
  \end{tabular}
  \caption{\textbf{Structured agent baseline} (F-score) on the 40 sketches of the minimal-prompt test, all with the same LLM and the same coordinates (\ours{}: coordinates and label, no views). LL3M-style vs.\ \ours{}: $p=0.18$ (better on 20 of 40); vs.\ trained: $p=0.24$; vs.\ minimal: $p<10^{-3}$.}
  \label{tab:ll3m}
\end{table}

\paragraph{The strict instruction on held-out sketches.}
We tested the strict fidelity sentence (\fidelityref) on 80 sketches after seeing the balanced results. We then ran it, unchanged, on the other 320 test sketches, none of which was used to develop the instruction, the prompt or the check (they had only been evaluated with the balanced default; \cref{tab:strict_heldout}). All 320 programs built. Strict scores 74.1, against 71.3 for balanced and 69.5 for the trained model on the same sketches; it is better than balanced on 243 of 320 ($p<10^{-19}$) and than the trained model on 214 ($p<10^{-8}$), with the largest gains on airplanes, chairs and tables, while cabinets still trail the trained model (54.4 vs.\ 61.5). After alignment, strict remains ahead of the trained model (76.2 vs.\ 74.7, $p=0.001$). On all 400 sketches strict reaches 74.5 against 69.8 (better on 268). The main results of the paper remain those of the balanced default.

\begin{table}[t]
  \centering
  \footnotesize
  \setlength{\tabcolsep}{2.4pt}
  \begin{tabular}{@{}lccccccc@{}}
    \toprule
    & \multicolumn{4}{c}{F-score (official)} & \multicolumn{3}{c}{F-score (aligned)} \\
    \cmidrule(lr){2-5}\cmidrule(l){6-8}
    & Strict & Balanced & Trained & S$>$T & Strict & Balanced & Trained \\
    \midrule
    Airplane & 94.8 & 93.1 & 87.1 & 66/80 & 95.6 & 94.3 & 92.8 \\
    Cabinet & 54.4 & 50.2 & 61.5 & 24/80 & 58.2 & 53.4 & 64.0 \\
    Chair & 77.8 & 73.8 & 69.5 & 62/80 & 79.3 & 76.1 & 74.8 \\
    Table & 69.5 & 68.1 & 60.0 & 62/80 & 71.6 & 70.4 & 67.2 \\
    \midrule
    All (320) & 74.1 & 71.3 & 69.5 & 214/320 & 76.2 & 73.5 & 74.7 \\
    \bottomrule
  \end{tabular}
  \caption{\textbf{Strict instruction on 320 held-out sketches}, the test sketches outside the 80 of the first strict run, not used to develop the instruction, the prompt or the check; the strict sentence is unchanged. S$>$T: sketches on which strict scores higher than the trained model. Strict vs.\ balanced $p<10^{-19}$, vs.\ trained $p<10^{-8}$; aligned, vs.\ trained $p=0.001$.}
  \label{tab:strict_heldout}
\end{table}

\paragraph{Repair with simpler acceptance scores.}
Simpler scores rank outputs about as well as ours on the 60 held-out sketches and better on all 380 balanced outputs not used for development, which include them (\cref{tab:check_design}). To test them end to end, we ran the repair loop on the 60 held-out balanced sketches with the same trigger, the same evidence and the same edit requests, but kept a round when the mean distance between strokes and surface in both directions fell by more than 1\%, or when the F-score against the strokes (\SI{2}{cm}) rose by more than 0.01 (\cref{tab:repair_alt}). Where a rule made the same decision as ours, the next request is the same, and we reused the stored candidate; where it differed, the chain continued from the model that rule kept, with new requests (1 and 2 sketches). Both rules give a gain similar to ours (74.4 and 74.3 vs.\ 73.8; $p=0.16$ and $0.92$). The gain thus does not depend on the particular acceptance score: accepting every repair also raises the mean F-score (\cref{tab:accept}). To separate the evidence from the effect of a second LLM pass, we also ran the loop with the same trigger, the same number of rounds and the same acceptance rule, but replaced the evidence by all stroke coordinates (as sent for generation) and the request to compare the script with them and fix the model. This control reaches 73.9 (22 sketches higher, 9 lower than without repair; $p=0.004$), against 73.8 for ours (18 higher, 1 lower; ours vs.\ control $p=0.53$). It keeps a repair for 23 of the 32 flagged sketches (ours 19), but a kept round changes a median of 6.5 of 11 parts (75\% of a model's parts), against 2 (25\%) with the evidence. The evidence thus does not raise the mean gain; it confines the repair to the parts that need it and avoids making sketches worse. The control cost \$15.8 for the 60 sketches.

\begin{table}[t]
  \centering
  \footnotesize
  \setlength{\tabcolsep}{3pt}
  \begin{tabular}{@{}lccc@{}}
    \toprule
    Acceptance score & F $\uparrow$ & + / -- & vs.\ ours \\
    \midrule
    No repair & 72.7 & & \\
    Ours (area score, coverage $-$0.01 at most) & 73.8 & 18 / 1 & -- \\
    Mean distance, both directions & 74.4 & 17 / 1 & $p=0.16$ \\
    F-score against the strokes & 74.3 & 13 / 2 & $p=0.92$ \\
    \bottomrule
  \end{tabular}
  \caption{\textbf{End-to-end repair with simpler scores} on the 60 held-out balanced sketches: the same trigger, evidence and edit requests, but a round is kept when the given score improves; when a decision differs from ours, the chain continues from the model it kept (new requests). ``+ / --'': sketches whose F-score repair raised / lowered.}
  \label{tab:repair_alt}
\end{table}

\paragraph{Editing a segmented mesh.}
For an edit that rescales one part, any segmented mesh can leave the other parts untouched. As a baseline we took our own generated parts as the segmentation, perfect by construction, and scaled the target part exactly as requested, either about its bounding-box center or about the points where it touches other parts (\cref{tab:mesh_edit}). On the 30 of the 32 rescaling edits whose target could be matched to strokes (those of \cref{tab:resketch}, so that both baselines are compared on the same edits) it leaves all other parts untouched, as our edits do, meets the requested size by construction (30 of 30; ours 27) and keeps 98 or 100 of 107 contacts with neighboring parts (ours 103). The locality of rescaling edits thus comes from the part structure. What distinguishes the program is that it supplies this structure and that it can carry out the other 27 edits, which make a part rounder, restyle it or add details and are not transforms; it does so with all 92 contacts kept, although added details are often too faint for the judges to see (\cref{tab:edit_judge}).

\begin{table}[t]
  \centering
  \footnotesize
  \setlength{\tabcolsep}{3pt}
  \resizebox{\linewidth}{!}{\begin{tabular}{@{}lccc@{}}
    \toprule
    & Size as asked & Contacts kept & Others changed \\
    \midrule
    \multicolumn{4}{@{}l}{\textit{30 of 32 rescaling edits (target matched to strokes)}} \\
    \ours{}: edit code & 27/30 & 103/107 & 0 \\
    Mesh, scale about center & 30/30 & 98/107 & 0 \\
    Mesh, scale about contacts & 30/30 & 100/107 & 0 \\
    \midrule
    \multicolumn{4}{@{}l}{\textit{27 other edits (rounder, restyle, details)}} \\
    \ours{}: edit code & 26/27 & 92/92 & 0 \\
    Mesh & \multicolumn{3}{c}{not a transform} \\
    \bottomrule
  \end{tabular}}
  \caption{\textbf{Editing a segmented mesh} (``Mesh''). The segmentation is our own generated parts, \ie perfect for this baseline; the target part is scaled exactly as requested, about its bounding-box center or about the points where it touches other parts. Same metrics as for our edits: bounding-box ratio within tolerance, contacts with neighboring parts (closer than 1\,cm) kept, parts other than the target that change by more than 1\,mm.}
  \label{tab:mesh_edit}
\end{table}

\ifsupp
\Cref{tab:resketch} gives the comparison with re-sketching summarized in \editingref.

\begin{table}[t]
  \centering
  \small
  \setlength{\tabcolsep}{3pt}
  \resizebox{\linewidth}{!}{\begin{tabular}{@{}lccc@{}}
    \toprule
    & \multicolumn{2}{c}{Outside surface moved} & Edits with \\
    & $>$1\% & $>$2\% & leakage \\
    \midrule
    \ours{}: edit the part's code & \textbf{0.0\%} & \textbf{0.0\%} & \textbf{0/30} \\
    Trained: re-sketch, same seed & 23.2\% & 12.7\% & 14/30 \\
    \quad after ICP alignment & 11.5\% & 6.1\% & 13/30 \\
    Trained: re-sketch, new seed & 32.4\% & 16.8\% & 30/30 \\
    \textit{Trained: no edit, new seed} & \textit{13.2\%} & \textit{3.6\%} & \textit{29/30} \\
    \bottomrule
  \end{tabular}}
  \caption{\textbf{Edit locality vs.\ re-sketching} on 30 rescaling edits: share of the surface outside the target region (its bounding box enlarged by 10\% of the object size) that moves by more than 1\% or 2\% of the object size; an edit ``leaks'' if more than 1\% of it moves by more than 1\%. We call an edit that does not leak ``ignored'' if, inside the target region as well, less than 1\% of the surface before or after the edit lies more than 1\% of the object size from the other surface (13 of 30 with the same seed).}
  \label{tab:resketch}
\end{table}
\fi

\paragraph{Rotation and missing components.}
A VR sketch is drawn where the object should be, so the output should follow it there. We rotated the 40 sketches of \cref{tab:ll3m} about the vertical axis by \ang{45} and \ang{90}, rotating the reference with them, and ran both methods unchanged (\cref{tab:robust}). Unlike in \cref{tab:ll3m}, \ours{} receives its default input: coordinates, views and label. At \ang{45} our F-score falls from 69.1 to 64.2 ($p=0.02$): the programs build such an object at an angle less accurately than an axis-aligned one. At \ang{90}, which only exchanges axes, it does not fall (70.7, $p=0.66$). The trained model generates in its canonical orientation and falls to 43.7 and 47.2 ($p<10^{-7}$ for both). Consistency, the F-score between the output for the rotated sketch, rotated back, and the output for the original sketch, is 75.0 and 79.8 for \ours{}, close to the 81.7 between two runs on the same unrotated sketch (39 sketches with a repeated run), and 44.4 and 53.8 for the trained model. To test incomplete sketches beyond random stroke dropout (\dropoutref), we removed every stroke of one component: each stroke was assigned to the part of our original output nearest to most of its points, and one part with at least three strokes and at most 40\% of them was chosen at random with a fixed seed, for example a wing, an engine, a chair leg, a side panel or an apron (12\% of the strokes on average). The reference keeps the component. This costs \ours{} 2.2 points (66.9, $p=0.52$) and the trained model 3.9 (65.1, $p=0.005$); on these sketches the two do not differ significantly ($p=0.53$).

\begin{table}[t]
  \centering
  \footnotesize
  \setlength{\tabcolsep}{4pt}
  \begin{tabular}{@{}lcccc@{}}
    \toprule
    & \multicolumn{2}{c}{F-score} & \multicolumn{2}{c}{Consistency} \\
    \cmidrule(lr){2-3}\cmidrule(l){4-5}
    Sketch & \ours{} & Trained & \ours{} & Trained \\
    \midrule
    Original & 69.1 & 69.0 & -- & -- \\
    Rotated \ang{45} & 64.2 & 43.7 & 75.0 & 44.4 \\
    Rotated \ang{90} & 70.7 & 47.2 & 79.8 & 53.8 \\
    One component removed & 66.9 & 65.1 & -- & -- \\
    \bottomrule
  \end{tabular}
  \caption{\textbf{Rotation and missing components} on the 40 sketches of \cref{tab:ll3m}, \ours{} with its default input of coordinates, views and label (\cref{tab:ll3m} uses coordinates and label only). Rotations are about the vertical axis; the reference is rotated with the sketch. One component removed: every stroke of one part (chosen at random among parts with at least three strokes and at most 40\% of them; 12\% of strokes on average), reference unchanged. Consistency: F-score between the output for the rotated sketch, rotated back, and the output for the original sketch. Ours vs.\ trained: \ang{45} $p<10^{-7}$, \ang{90} $p<10^{-7}$, component removed $p=0.53$.}
  \label{tab:robust}
\end{table}

\paragraph{Proportions.}
As a measure of geometric conditioning beyond the F-score, we compared the proportions of each output's bounding box (1st to 99th percentile of surface points along each axis, as log ratios) with those of its reference on all 400 sketches. After removing each method's category mean, the outputs follow the deviations of the individual references with a slope of 0.99 for \ours{} and 0.93 for the trained model, which pulls proportions slightly toward its category mean, most on airplanes (0.91); the median proportion error is 1.3\% vs.\ 1.7\% (\ours{} lower on 252 of 400, $p<10^{-7}$). Both follow the overall proportions closely; the difference is small.

\section{Best and worst cases}
\label{app:qual}
\begin{figure*}[p]
  \centering
  \includegraphics[width=0.49\linewidth]{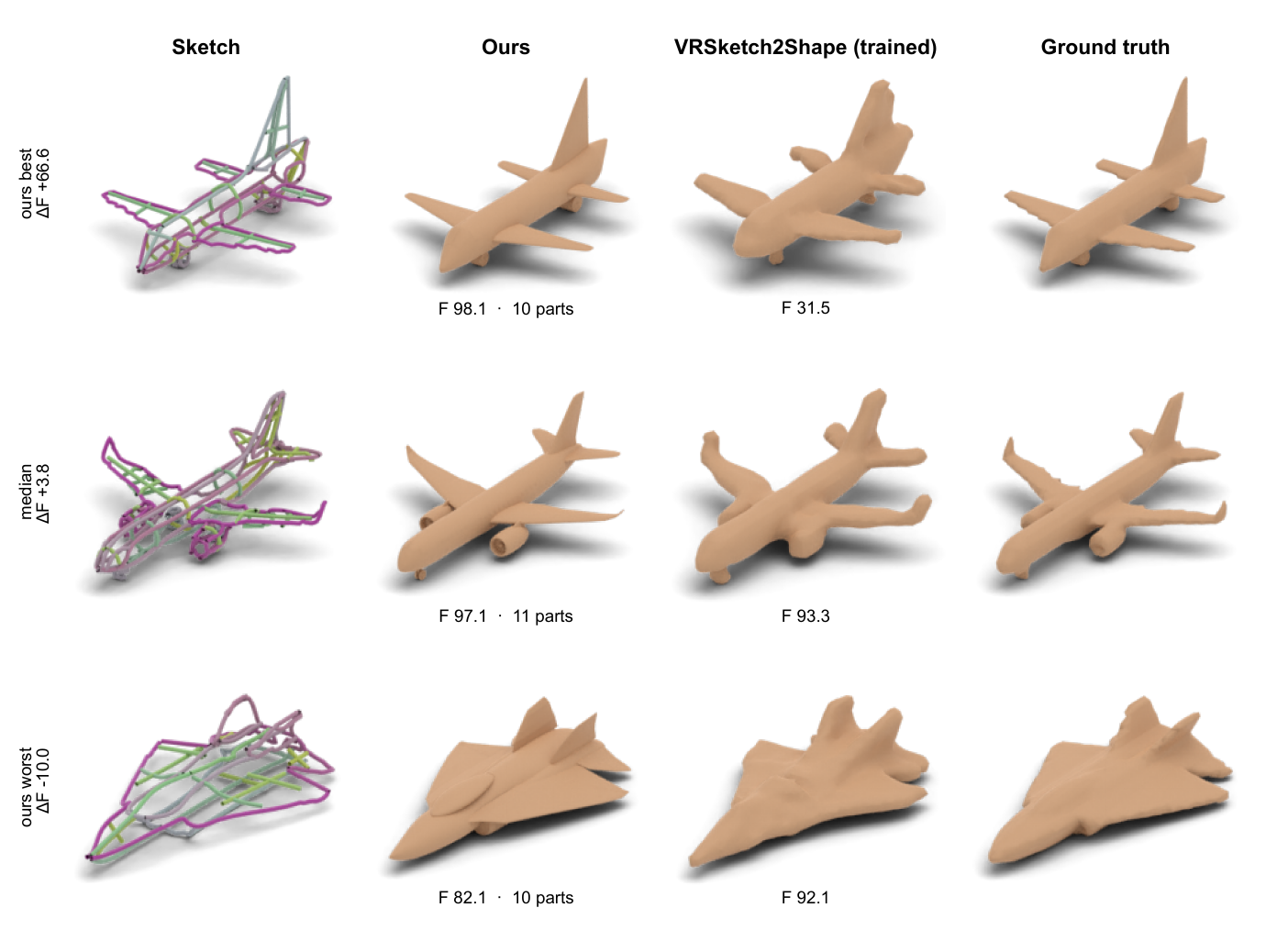}\hfill
  \includegraphics[width=0.49\linewidth]{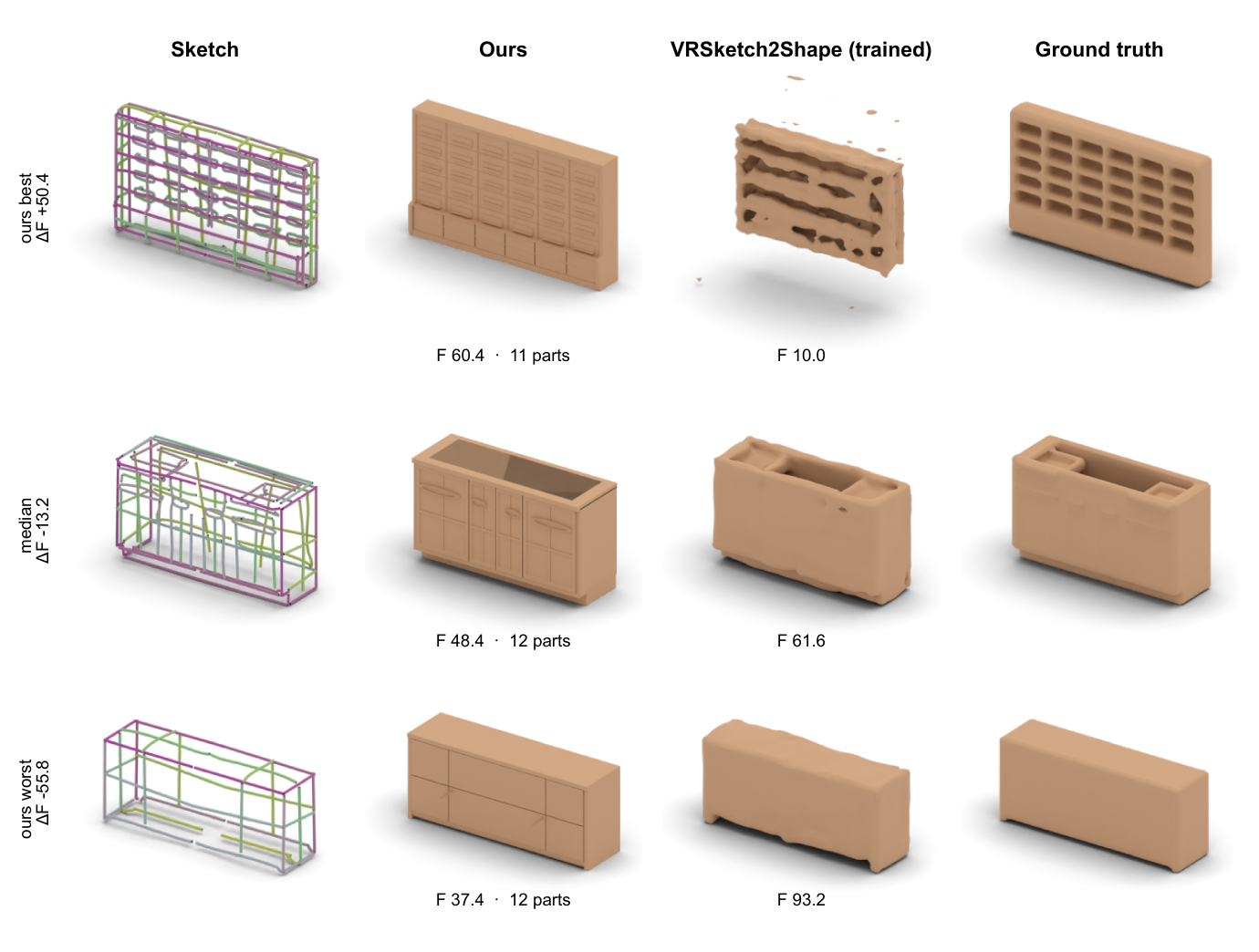}\\[1ex]
  \includegraphics[width=0.49\linewidth]{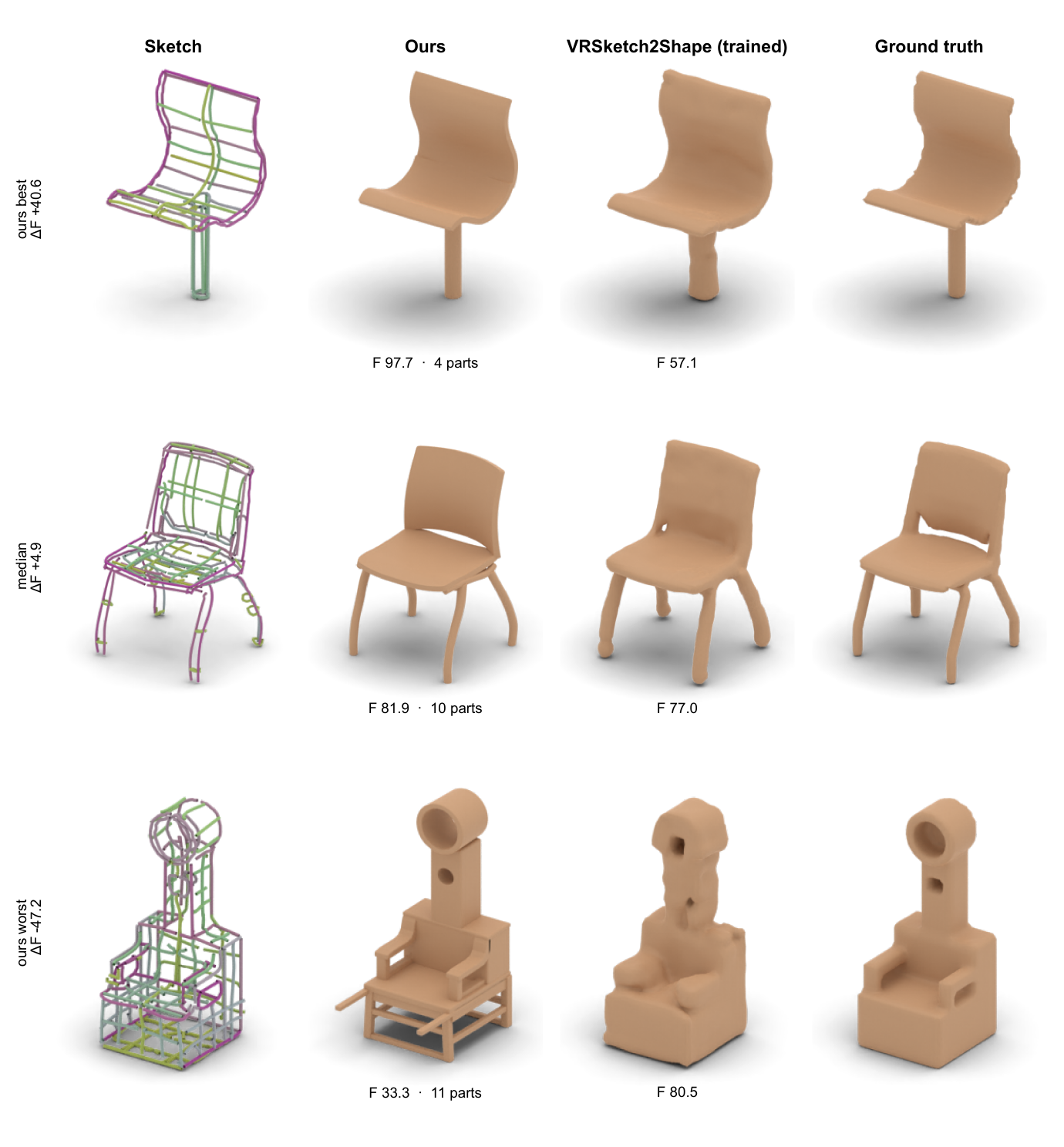}\hfill
  \includegraphics[width=0.49\linewidth]{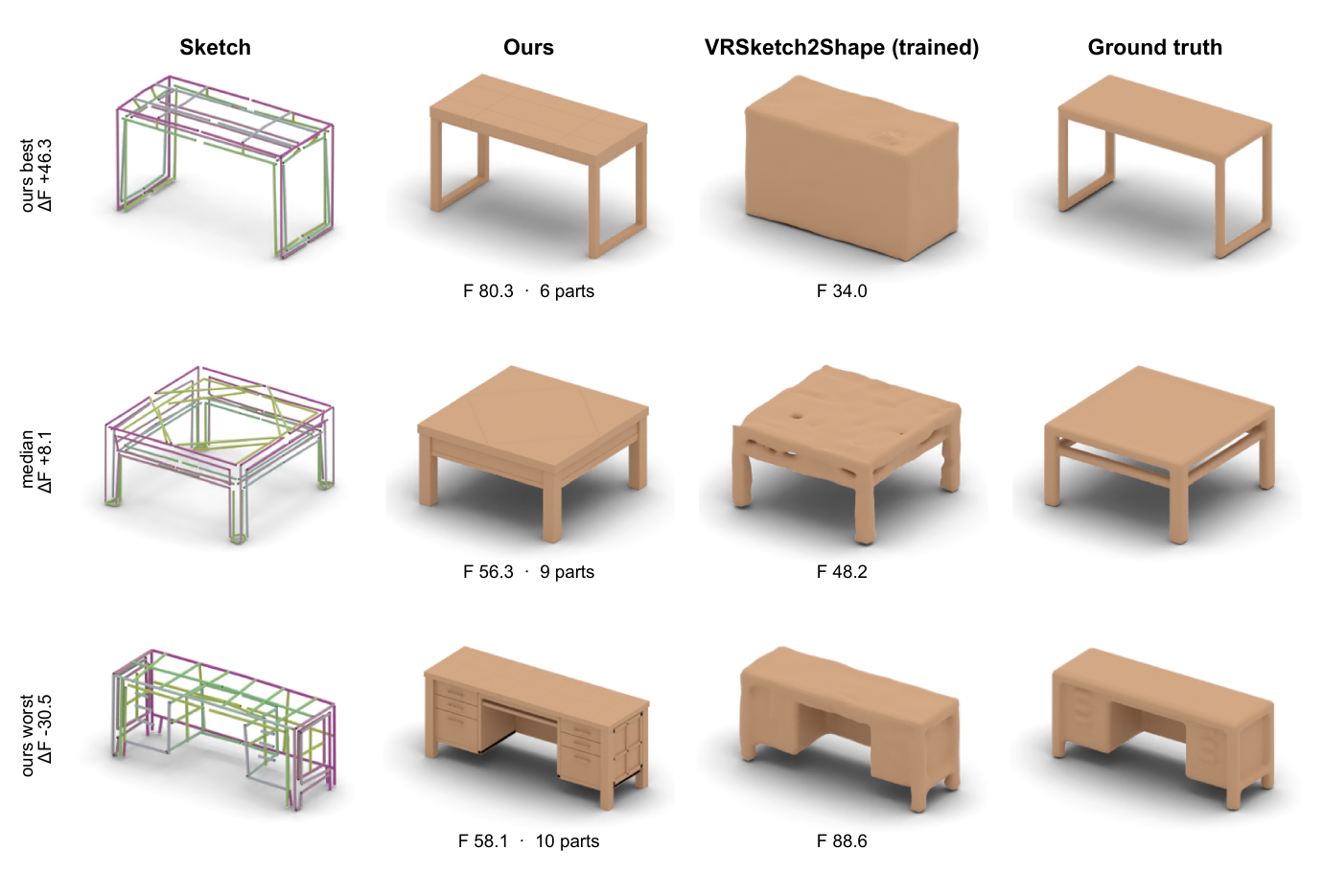}
  \caption{\textbf{Largest gain, median and largest loss per category} (airplane, cabinet; chair, table). Columns: sketch, \ours{}, trained VRSketch2Shape, ground truth.}
  \label{fig:qual_all}
\end{figure*}

\Cref{fig:qual_all} shows, for each category, the sketches with the largest gain, the median, and the largest loss of \ours{} relative to the trained model in F-score.

\section{Main methods side by side}
\label{app:routes}
\begin{figure*}[t]
  \centering
  \includegraphics[width=\linewidth]{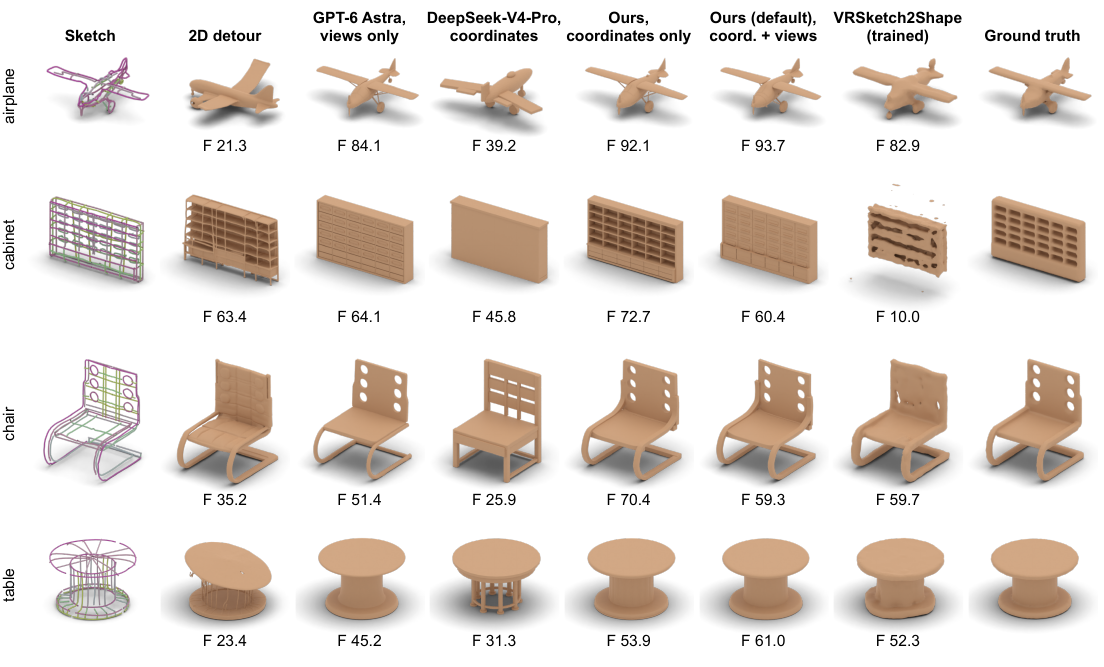}
  \caption{\textbf{Main methods on the same sketches}, one per category: the sketch at the median F-score of \ours{} (default input) among the 18 sketches on which every method was run. ``Coordinates only'' (ours) and DeepSeek-V4-Pro receive the same prompt and input; GPT-6 Astra with views only receives the five renderings instead of the coordinates. The 2D detour is shown at its best rotation. All shapes in the same color, each normalized to its own bounding box as in the evaluation; strokes are colored by drawing order, from purple (first) to yellow (last).}
  \label{fig:routes}
\end{figure*}

\Cref{fig:routes} shows the outputs of the main methods on the same sketches, complementing \cref{tab:routes}. All of them were run on 18 sketches (the intersection of the subsets in \cref{tab:routes}); for each category we show the one at the median F-score of \ours{} with its default input, so the examples are not selected for appearance. Luo \etal and LAS-Diffusion are not shown: their numbers are those reported by~\cite{chen2026vrsketch2shape}, and their outputs are not available. The minimal prompt, the LL3M-style agent, Claude Fable 5.2 and Gemini 4 Argon were also run on these sketches but are not shown. In these examples, the 2D detour replaces the drawn airplane by a different aircraft and turns the hatched table base into a cage of thin rods; DeepSeek-V4-Pro builds a four-legged chair instead of the drawn cantilever frame and a ring of columns instead of the drawn drum base; the trained model's cabinet has floating fragments, which cost it most of its F-score.

\section{Edit instructions}
\label{app:edits}
Simple edits (20 models drawn at random from the first 40 generated models; target part and instruction drawn uniformly at random; one edit was excluded because our part extraction had named a child object as its target, leaving 19; counted as a failure, it would make 55 of 60 single-part edits meet the size check instead of 55 of 59): ``Make this part about 30\% thicker.'', ``Make this part about 20\% longer.'', ``Give this part a more rounded, softer shape.'', ``Make this part about 25\% wider.''

Hard edits (40 models not used for the simple edits, 10 per category, target part chosen at random, kinds assigned in rotation):
\begin{itemize}
  \item restyle: ``Redesign this part in a more ornate, curved style, keeping its overall size and where it sits.''
  \item detail: ``Add small decorative details to this part (grooves, trim or rivets), keeping its overall size.''
  \item bigger: ``Make this part about 40\% bigger in every direction, and keep it attached where it is attached now.''
  \item thinner: ``Make this part about half as thick (more slender and delicate), keeping its length and where it sits.''
\end{itemize}
Three-step chains (10 further models with at least five parts, used in neither set, chosen in rotation over the categories): each chain edits three different parts drawn at random, with three different instructions drawn from thicker, longer, wider, restyle, bigger and thinner; each edit is applied to the program returned by the previous one, and every step is compared with the step before it, built with the same builder.

An edit counts as following the requested size if the target part's bounding-box ratios fall in these ranges: thicker, largest ratio 1.15--1.48; longer, 1.10--1.32; wider, 1.12--1.40 (in each case all other ratios $>0.9$); rounded, all ratios within $\pm10\%$; restyle, within $\pm15\%$; detail, 0.9--1.12; bigger, median ratio 1.2--1.6 and all $>1.1$; thinner, smallest ratio 0.3--0.75 with the longest axis within $\pm10\%$. These ranges test the size of the change but not its axis, so we also checked the axes of the passing single-part edits (not those of the three-step chains). In each of the 11 thicker, longer and wider edits, only the requested direction changed (all other axes by less than 2\%): longer edits lengthened the part's longest axis (for a chair back of 25.8 by \SI{25.0}{cm}, the vertical one), thicker edits enlarged its shorter axes, and wider edits enlarged it along the object's left--right axis. All 10 ``bigger'' edits scaled the part by exactly 1.4 along every axis, and 6 of the 8 passing ``half as thick'' edits halved the part's thinnest axis; the other 2, a drawer and a drawer bank, halved its middle one.

\section{Edit judgments}
\label{app:edit_judge}
\begin{figure}[t]
  \centering
  \includegraphics[width=0.62\linewidth]{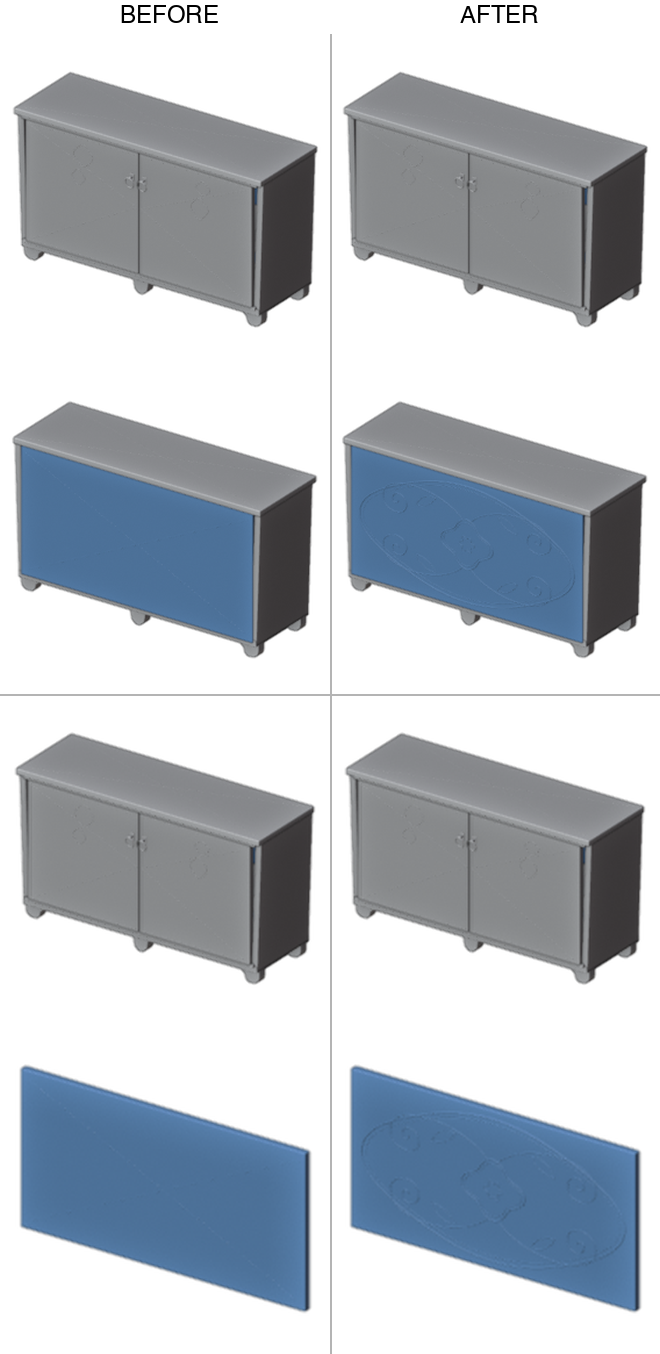}
  \caption{\textbf{Input to the edit judges} for one restyle edit (``Redesign this part in a more ornate, curved style, keeping its overall size and where it sits.''; target: the back panel of a cabinet). Both judges answered ``yes''; Gemini 4 Argon: ``The back panel gained an ornate oval medallion with curved relief decorations while keeping the same size and position; the rest of the cabinet appears unchanged.''}
  \label{fig:edit_judge_example}
\end{figure}

Each of the 59 single-part edits was rendered before and after in the same framing (\cref{fig:edit_judge_example}): the whole object from the front right and from the back left, a close-up of the target part in place (front right, framed on the target's bounding box), and the target part alone (back left). The target part is blue and all other parts are gray. The two judges, Claude Fable 5.2 (\texttt{anthropic/claude-fable-5.2}) and Gemini 4 Argon (\texttt{google/gemini-4-argon}), were queried through OpenRouter on November~14, 2026, with low reasoning effort and default sampling, one request per image, without retries or selection. The system message was:
\begin{quote}\small\ttfamily
You are evaluating a 3D modeling tool. You will see one image with renders of a 3D object: the left column is BEFORE an edit and the right column is AFTER it. Rows 1 and 2 show the whole object from two viewpoints (front-right and back-left); row 3 is a close-up of the edited part in place (front-right), and row 4 shows the edited part alone, without the rest of the object (back-left). The part the user asked to edit is shown in blue; all other parts are gray. In each row, BEFORE and AFTER use the same scale and framing, so size changes are visible.
\end{quote}
and the user message, with the image:
\begin{quote}\small\ttfamily
The edit instruction for the blue part (named "\{target\}") was: "\{instruction\}"\\
Answer three questions:\\
1. done: Was the instruction carried out on the blue part? "yes" (clearly), "partly" (only some of it, or barely visible), or "no" (not done, or the opposite was done).\\
2. others\_changed: Does any gray part look different between BEFORE and AFTER? "yes" or "no".\\
3. reason: one short sentence.\\
Reply with only a JSON object: \{"done": "...", "others\_changed": "...", "reason": "..."\}
\end{quote}
In the first control, the instruction is replaced by one that was not carried out: ``Make this part about 40\% smaller.'' for ``bigger'', the 20\%, 25\% and 30\% shrinking counterparts for ``longer'', ``wider'' and ``thicker'', and ``Make this part about 40\% bigger.'' for ``thinner'' (the opposite size change) and for ``rounded'', ``restyle'' and ``detail'', which keep the size. In the second control, the before image is shown on both sides with the original instruction. All 354 replies were valid JSON. The judgments cost \$2.70 (Claude Fable 5.2) and \$0.57 (Gemini 4 Argon) in total, controls included.

\section{Other LLMs}
\label{app:openllm}
We queried DeepSeek-V4-Pro (model \texttt{deepseek/deepseek-v4-pro-0813}) through OpenRouter on November~11, 2026, and Claude Fable 5.2 (\texttt{anthropic/claude-fable-5.2}) and Gemini 4 Argon (\texttt{google/gemini-4-argon}) on November~14, 2026, with the system prompt and user message of \cref{app:prompts} (coordinates and category name, no views), default sampling and reasoning settings and an output limit of 32{,}000 tokens, which includes the model's reasoning. No output limit was set for GPT-6 Astra (default settings); on these 60 sketches it used at most 12{,}875 output tokens per sketch, reasoning and fix round included, so the same limit would not have cut off any of its replies. The returned scripts were built with the same Blender~3.6 builder, with the same single fix round after a build error or a single-part result, and scored with the same protocol. The 60 sketches are the first 15 per category, in a fixed interleaved order, of the 100 in \cref{tab:ablation}; every model ran on all 60, and every comparison uses the same sketches for both models. A reply without a Python code block counts as a failure with an F-score of zero; the 11 failures of DeepSeek-V4-Pro (7) and Claude Fable 5.2 (4) were replies that reached the output limit, and the one of Gemini 4 Argon a script that still failed to build after the fix round. By category, DeepSeek-V4-Pro reaches 31.6, 47.7, 30.3 and 52.6 on airplanes, cabinets, chairs and tables, against 87.3, 59.2, 65.1 and 72.3 for GPT-6 Astra; Claude Fable 5.2 76.2, 63.6, 60.1 and 73.2; Gemini 4 Argon 83.2, 59.1, 59.4 and 71.0. Gemini 4 Argon needed the fix round on 15 of 60 sketches, Claude Fable 5.2 on none. The programs have a median of 9 (DeepSeek-V4-Pro, Gemini 4 Argon) and 10 (Claude Fable 5.2) named parts. Per sketch, the queries cost \$0.07 (DeepSeek-V4-Pro), \$0.12 (Gemini 4 Argon) and \$0.80 (Claude Fable 5.2) on average, against \$0.31 for GPT-6 Astra.

\section{2D-detour baseline}
\label{app:detour}
\begin{figure*}[t]
  \centering
  \includegraphics[width=0.8\linewidth]{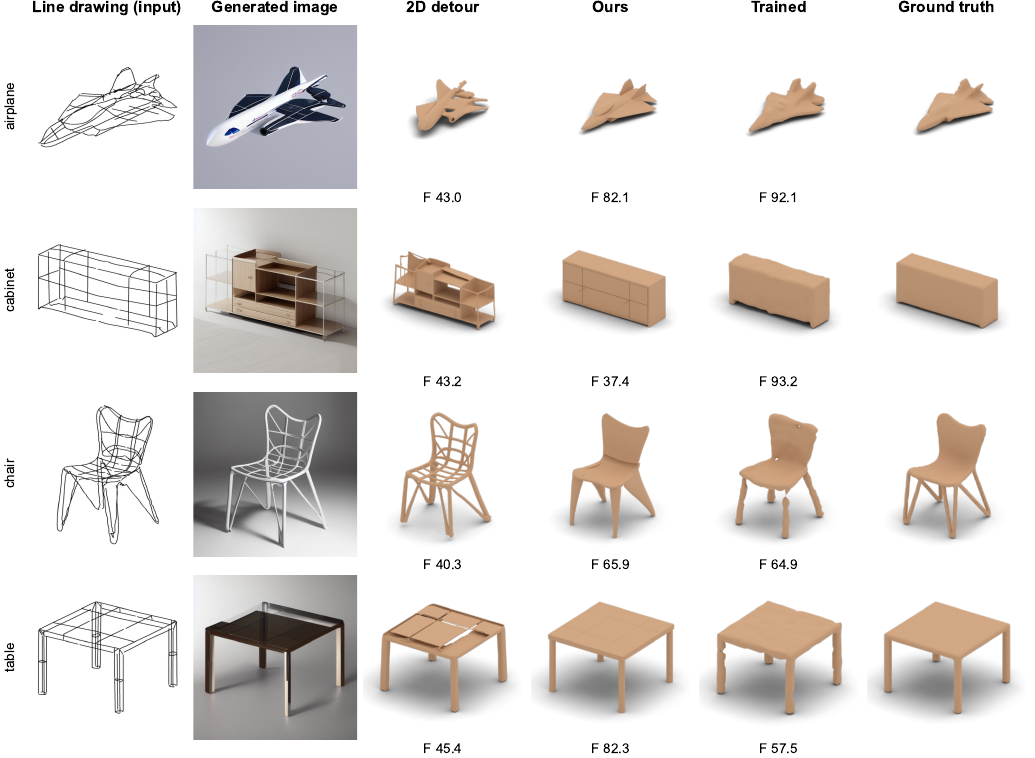}
  \caption{\textbf{2D detour vs.\ \ours{}}, one sketch per category: the sketch whose detour F-score is closest to the category mean. The line drawing is the detour's input; the generated image is the ControlNet output that Hunyuan3D-2 lifts to 3D. Meshes are shown after the evaluation's normalization, the detour at its best rotation. Hidden edges and hatching strokes, indistinguishable from structure in one drawing, become wires, glass panes and frames.}
  \label{fig:detour}
\end{figure*}

\Cref{tab:detour_cats} gives the results of the 2D detour (\cref{sec:detour}) by category, and \cref{fig:detour} shows one sketch per category.

\begin{table}[t]
  \centering
  \footnotesize
  \setlength{\tabcolsep}{3pt}
  \begin{tabular}{@{}lccccc@{}}
    \toprule
    & Air. & Cab. & Chair & Table & All \\
    \midrule
    2D detour, best rotation & 43.2 & 43.5 & 39.9 & 46.8 & 43.3 \\
    2D detour, one rotation & 36.5 & 42.1 & 39.2 & 45.4 & 40.8 \\
    \ours{}, coordinates only & 88.7 & 58.2 & 66.4 & 71.7 & 71.3 \\
    \ours{}, coord.\ + views & \textbf{89.8} & 56.8 & \textbf{67.3} & \textbf{72.3} & \textbf{71.6} \\
    \textit{Trained} & \textit{85.4} & \textit{\textbf{69.5}} & \textit{66.5} & \textit{63.0} & \textit{71.1} \\
    \bottomrule
  \end{tabular}
  \caption{\textbf{2D detour by category} (F-score) on the 100 sketches of \cref{tab:ablation} (25 per category). Best rotation: per sketch; one rotation: the same for all sketches. \ours{} with coordinates and views is the default. Median Chamfer distance: 13.33 for the detour, 2.81 for \ours{} with coordinates only.}
  \label{tab:detour_cats}
\end{table}

\paragraph{Pipeline.} The strokes are projected orthographically from \ang{35} azimuth and \ang{20} elevation (front left, above), scaled to fill a $512\times512$ image with an 8\% margin, and drawn as black polylines on white, far strokes first. The drawing, inverted and dilated by one pixel as the scribble model expects, conditions the \texttt{control\_v11p\_sd15\_scribble} ControlNet~\cite{zhang2023controlnet} on Stable Diffusion~1.5~\cite{rombach2022ldm} (25 UniPC steps, guidance 7.5). Prompt: ``a \{category\}, a single \{category\} on a plain white background, product photo, studio lighting, whole object in view, high quality''; negative prompt: ``cropped, out of frame, multiple objects, people, text, watermark, cluttered background, lowres, blurry, deformed'', without ``people'' for the person sketches. The background is removed with the remover bundled with Hunyuan3D-2, and Hunyuan3D-2~\cite{zhao2025hunyuan3d} (shape model v2.0, half precision) generates the mesh with 30 steps at octree resolution 256. Each sketch has a fixed seed. A sketch takes 5 to 7 minutes on an Apple M2 Max laptop (about 1 minute for the image, the rest for the shape).

\paragraph{Evaluation.} The mesh is rotated about the vertical axis by each multiple of \ang{15}; each rotated mesh is normalized and scored with the protocol of \cref{sec:setup}, and the rotation with the best F-score is kept. The best single rotation for all 100 sketches is \ang{0} (mean F-score 40.8); per sketch, 76 of the 100 best rotations are \ang{0} or \ang{180}, as expected for objects that are nearly symmetric front to back. Nine of the 25 airplanes come out turned by about \ang{90}, so a single rotation lowers the airplane score most (36.5 instead of 43.2).

\onecolumn
\section{Prompts}
\label{app:prompts}
The prompts below are used verbatim; long lines are wrapped for display. Placeholders are in braces. The views sentence is omitted when no views are attached, and the stroke data is omitted when only views are attached.

{\lstset{basicstyle=\ttfamily\footnotesize, upquote=true}%
\par\medskip\noindent\textbf{Generation: system prompt.}\par\nopagebreak\smallskip
\begin{lstlisting}
You are a 3D artist who turns a user's rough mid-air 3D sketch into a clean, realistic model by
  writing a Blender script. Follow the shape-fidelity level given in the request. Keep every
  attribute the user stated (colors, materials, counts) exactly.
Blender 3.6 Python (bpy) rules:
- Units are meters. Z is up. Coordinates match the user's stroke data: X right, Y back, Z up,
  origin = bottom center of the sketch, so the object's base sits at z=0. The front of the
  object faces -Y.
- Build the object as SEPARATE mesh objects, one per meaningful editable part (e.g. Roof,
  Chimney, FrontDoor, Walls, Wheels). Name every top-level part object with a short unique
  PascalCase English name (letters and digits only). Small details that belong to a part (e.g.
  a door handle) must be parented to that part's object. Give every object, parented details
  included, its final world-space position: parenting keeps an object where it is. Use 3 to 12
  top-level parts. Never join parts together.
- Assemble it as one object: every part touches or is attached to the part it belongs on (no
  floating parts, no gaps). Doors, windows, shutters, hatches and lids are CLOSED and flush
  with (or set slightly into) their wall or surface unless the user asks for them open. Nothing
  may stick out beyond the sketch's outline by more than about 2 cm unless the strokes show it.
- Clean, smooth geometry (bevels, subdivision, screw/lathe for round parts). Apply all
  modifiers at the end.
- If the object is a person, animal, creature or character (anything that could move), build it
  ready to be given a skeleton and animated: (a) each arm, leg, neck, tail and long ear is ONE
  continuous tube or loft along the limb's centre line with many cross-sections (at least 12
  along an arm or leg, closer together around the elbow and knee) so it can bend smoothly;
  limbs sink 1-2 cm into the body instead of just touching it; on a person, where skin is
  exposed, a rounded joint cap (sphere) sits at the elbows, wrists, knees and ankles (never
  under clothes, never poking out of them; never on animals). Give the body its natural depth
  even if the strokes are drawn almost flat (a person's chest is about 60% as deep as it is
  wide). Clothes follow the body with a little thickness: a shirt is a rounded torso volume
  (chest, back, shoulders); sleeves, shorts legs and trouser legs are separate snug tubes
  wrapped around the limb so they bend with it, starting inside the shirt's armhole or the
  shorts' leg opening, in the same colour, so the garment still reads as one piece. (b) A
  person's or creature's hand has a separate thumb and 3 or 4 fingers, slightly spread, each
  finger built from 3 short segments that are separate child objects (names like IndexL1,
  IndexL2, IndexL3, ThumbR1) with a small sphere at each knuckle; animal paws need no fingers.
  (c) A mouth that can open without changing the head: NEVER split or cut the head into upper
  and lower halves. On the face build, as child objects of the head part: MouthOpening, a dark
  (deep red or near black) flattened ellipsoid about a third of the face wide, centred just
  under the face surface so only a thin dark line shows when closed; UpperLip and LowerLip,
  thin curved tubes along its top and bottom edges; teeth or a tongue inside it if wanted.
  Opening the mouth makes the opening taller and moves the lower lip down; the head never
  moves. An animal with a long snout may instead have a separate lower jaw (Jaw) hinged at the
  back of the mouth. (d) Eyes that can close: each eye is a sphere, and its eyelid is a
  separate child object (EyelidL, EyelidR): a thin spherical shell slightly larger than the
  whole eye (iris and pupil included, so nothing pokes through when it closes) that wraps from
  behind the top of the eye forward to just above the pupil (about three quarters of the upper
  half, back to front), hinged at the eyeball's centre, so rotating it down closes the eye
  while the top of the eye stays covered. Brows are separate child objects. (e) Animals and
  other organic creatures (not people or robots) must look like a soft sculpted toy, not
  stacked primitives: read box and straight-line strokes only as proportions. Build the torso,
  legs, neck and tail as ONE smooth Body from metaballs, then convert it to a mesh: ELLIPSOID
  elements for the chest, belly and rump (chest deeper than the waist), a thigh or shoulder
  ellipsoid where each leg meets the body, and ball chains along each leg (thick at the top,
  tapering, ending in a slightly wider rounded paw or hoof), along the neck and along the tail
  (tapering to the tip). Metaball settings: threshold 0.6, every element stiffness 2.0,
  resolution about 3.5 mm; a BALL's visible radius is about 2/3 of its radius, so set radius =
  1.5 x the wanted thickness; an ELLIPSOID gets radius 0.015 and size_x/size_y/size_z = the
  wanted half-sizes in centimetres; space the balls of a chain no more than 0.3 x their radius
  apart (wider spacing looks like a string of beads); then
  bpy.ops.object.convert(target='MESH') and give it smooth shading. The head is a separate
  rounded part with a distinct muzzle or snout that sinks into the end of the neck. No joint
  spheres on fur-covered legs. Fur markings (stripes, spots, a lighter belly) lie flat on the
  fur: duplicate the Body mesh, move every vertex 0.6 mm out along its normal, and Boolean-
  INTERSECT it (EXACT solver) with a cutter object made of one simple slab or blob per marking
  (cutter shapes must not overlap each other), keep the result as a child of Body in the
  marking's colour and delete the cutter; never thin wire rings or loose strips.
Keep all of these as children of their part (fingers under the arm or hand part, Jaw, eyes and
  eyelids under the head part), so the top-level parts stay the same.
- Keep the model light (it is shared between headsets): small details such as knuckles,
  fingertips, joint caps, eyes, buttons and bolts use low-resolution spheres and cylinders
  (about 12 segments, no subdivision). Aim for under 50,000 vertices in total.
- Materials: Principled BSDF with constant values only (Base Color, Metallic, Roughness,
  Transmission, IOR, Alpha). No image textures and no procedural texture nodes.
- Do NOT add cameras or lights, do not render, do not save files.
- Reply with ONLY one ```python code block containing the full script.
\end{lstlisting}
\par\medskip\noindent\textbf{Generation: user message.}\par\nopagebreak\smallskip
\begin{lstlisting}
The user drew a rough 3D sketch in mixed reality. Images: orthographic front, left, back,
  right, top views (blue lines; wireframe, so hidden lines are visible). The user said: "a
  chair". Sketch size: about {W} cm wide, {H} cm tall, {D} cm deep.
Exact 3D stroke data (cm; X right, Y back, Z up; origin = bottom center of the sketch). Strokes
  are outlines/edges of the object:
[{"id":1,"points_cm":[[...],...]},...]
Shape fidelity: BALANCED. Follow the strokes' overall size, proportions and where each part
  sits closely, but clean up and refine the shapes so the object looks well made.
Write the Blender script.
\end{lstlisting}
\par\medskip\noindent\textbf{Editing: system prompt.}\par\nopagebreak\smallskip
\begin{lstlisting}
You edit an existing Blender script that builds a 3D model made of named part objects. Apply
  ONLY the user's requested change. Every part you are not asked to change must stay byte-for-
  byte identical in the code (same geometry, materials and object names). Keep the edited
  part's object name unchanged.
<the same Blender rules as in the generation system prompt>
\end{lstlisting}
\par\medskip\noindent\textbf{Editing: user message.}\par\nopagebreak\smallskip
\begin{lstlisting}
Model parts: {comma-separated part names}.
Change ONLY the part object named "{part}" (and its child details). Edit request from the user:
  "{instruction}"
Shape fidelity: BALANCED. Follow the strokes' overall size, proportions and where each part
  sits closely, but clean up and refine the shapes so the object looks well made.
```python
{current script}```
\end{lstlisting}
\par\medskip\noindent\textbf{Shape-fidelity sentence, strict variant (replaces the BALANCED sentence above; \cref{sec:fidelity}).}\par\nopagebreak\smallskip
\begin{lstlisting}
Shape fidelity: STRICT. Match the strokes as exactly as you can: positions, proportions and
  outlines within about 1 cm. Do not add, remove or reshape anything the strokes do not show;
  the strokes take priority over typical proportions of the object category.
\end{lstlisting}
\par\medskip\noindent\textbf{Check against the sketch (\cref{sec:verify}): an actual message, sent as the edit request with no target part (balanced output for a development cabinet sketch). When strokes are missed, they are listed before the part table, each with its length, median distance, nearest part and coordinates.}\par\nopagebreak\smallskip
\begin{lstlisting}
Automatic check of this model against the user's 3D sketch. Coordinates are in cm in the same
  frame as the script (X right, Y back, Z up; divide by 100 for meters). Each stroke is an
  outline or edge the user drew, so the model surface should pass along the strokes, and the
  model should not contain parts the user did not draw.
Stroke support per part (share of the part's surface that is within 3 cm of a stroke or
  outlined by strokes lying in its own plane; total length of the strokes lying on the part;
  share of the model surface), least supported first:
- "Shelf": 29% near strokes, 0 cm of strokes on it, 12% of the surface
- "LeftSide": 67% near strokes, 110 cm of strokes on it, 7% of the surface
- "RightSide": 68% near strokes, 108 cm of strokes on it, 7% of the surface
- "CenterStile": 68% near strokes, 14 cm of strokes on it, 1% of the surface
- "RightDoor": 93% near strokes, 162 cm of strokes on it, 10% of the surface
- "LeftDoor": 93% near strokes, 157 cm of strokes on it, 10% of the surface
- "Back": 96% near strokes, 274 cm of strokes on it, 20% of the surface
- "Base": 98% near strokes, 413 cm of strokes on it, 17% of the surface
- "Top": 99% near strokes, 452 cm of strokes on it, 17% of the surface
Using this evidence and what the object is, fix the model so it matches the sketch: move,
  extend or reshape parts so their surfaces follow the strokes they miss, and delete (or shrink
  to what the strokes show) parts the user evidently did not draw. Change only the parts the
  evidence shows to be wrong; every other part must stay byte-for-byte identical.
\end{lstlisting}
\par\medskip\noindent\textbf{Minimal-prompt baseline (\cref{sec:ablation}): the entire request; no system prompt. Same data as the coordinates-only condition; only the instructions differ.}\par\nopagebreak\smallskip
\begin{lstlisting}
A user drew a rough 3D sketch in mixed reality. The user said: "a chair". Sketch size: about
  {W} cm wide, {H} cm tall, {D} cm deep.
3D stroke data (cm; X right, Y back, Z up; origin = bottom center of the sketch):
[{"id":1,"points_cm":[[...],...]},...]
Write a Blender 3.6 Python script that builds this object. Put the script in a ```python code
  block.
\end{lstlisting}
}

\end{document}